\documentclass[final,5p,times,twocolumn]{elsarticle}

\usepackage{hyperref}

\usepackage{subfig}
\usepackage{multirow}
\usepackage[fleqn]{amsmath}
\usepackage{booktabs}
\usepackage{algorithmic}
\usepackage{enumerate}
\usepackage{array}
\usepackage{gensymb}
\usepackage{stfloats}
\fnbelowfloat
\usepackage{url}
\usepackage{color}
\usepackage{stfloats}
\usepackage{amssymb}
\usepackage{arydshln}
\usepackage{stfloats}     
\usepackage{graphicx}     
\usepackage{multirow}     
\usepackage{array, caption, threeparttable}
\usepackage[font=small,labelfont=bf,labelsep=none]{caption}
\usepackage[marginal]{footmisc}

\usepackage{enumitem}
\usepackage{siunitx}

\usepackage{lineno}

\usepackage{listings}
\usepackage{xcolor}
\usepackage{float}

\usepackage{longtable}

\usepackage{graphicx}  
\usepackage{booktabs}  
\usepackage{makecell}  
\usepackage{multirow}

\usepackage[ruled,vlined]{algorithm2e}

\begin{document}

	\begin{frontmatter}

            \title{Unleashing the Capabilities of Large Vision-Language Models \\ for Intelligent Perception of Roadside Infrastructure}

		\author[liesmars]{Luxuan Fu\fnref{cofirst}}

            \author[liesmars]{Chong Liu\fnref{cofirst}}

            \author[liesmars]{Bisheng Yang\corref{correspondingauthor}}

            \author[liesmars,luojia]{Zhen Dong}
		
		\cortext[correspondingauthor]{Corresponding authors.}
		\address[liesmars]{State Key Laboratory of Information Engineering in Surveying, Mapping and Remote Sensing (LIESMARS), Wuhan University, Wuhan 430079, China}
        \address[luojia]{Hubei Luojia Laboratory, Wuhan 430079, China}
            \fntext[cofirst]{These authors contributed equally to this work.}

		\begin{abstract}

            Automated perception of urban roadside infrastructure is crucial for smart city management, yet general-purpose models often struggle to capture the necessary fine-grained attributes and domain rules. While Large Vision–Language Models (VLMs) excel at open-world recognition, they often struggle to accurately interpret complex facility states in compliance with engineering standards, leading to unreliable performance in real-world applications. To address this, we propose a domain-adapted framework that transforms VLMs into specialized agents for intelligent infrastructure analysis. Our approach integrates a data-efficient fine-tuning strategy with a knowledge-grounded reasoning mechanism. Specifically, we leverage open-vocabulary fine-tuning on Grounding DINO to robustly localize diverse assets with minimal supervision, followed by LoRA-based adaptation on Qwen-VL for deep semantic attribute reasoning. To mitigate hallucinations and enforce professional compliance, we introduce a dual-modality Retrieval-Augmented Generation (RAG) module that dynamically retrieves authoritative industry standards and visual exemplars during inference. Evaluated on a comprehensive new dataset of urban roadside scenes, our framework achieves a detection performance of 58.9 mAP and an attribute recognition accuracy of 95.5\%, demonstrating a robust solution for intelligent infrastructure monitoring.
            
            \end{abstract}


	\end{frontmatter}

    \section{Introduction}            
    


        As urban development shifts from rapid expansion to refined management, the intelligent perception of roadside infrastructure has become a critical priority~\cite{ma2022road_cv}. This task entails not only accurately detecting diverse facilities—such as traffic lights and signs—but also comprehensively interpreting their fine-grained attributes and physical conditions. However, manual inspection is prohibitively expensive, necessitating automated solutions that are not only accurate but also adaptable enough to handle this heterogeneity without imposing unmanageable annotation costs.
        
        While standard Computer Vision (CV) has standardized asset localization, it remains bound by a rigid, data-hungry paradigm. Traditional detectors typically rely on exhaustive supervision to establish feature representations from scratch, imposing a heavy data burden to achieve robust performance on domain-specific assets~\cite{bai2022infrastructure}. Moreover, without the capability for semantic reasoning, these methods struggle to capture fine-grained attributes—such as distinguishing specific damage conditions—creating a disconnect between simple bounding boxes and the rich, actionable insights required for intelligent maintenance~\cite{campbell2019detecting}.

        Recent advances in Vision–Language Models (VLMs) and Large Language Models (LLMs) offer a promising avenue for transcending these limitations by integrating visual recognition with natural language reasoning. These multimodal systems demonstrate exceptional capabilities in open-world perception and context-aware understanding~\cite{cui2024survey}. However, a critical gap remains between their descriptive prowess and the practical requirements of urban infrastructure management. Existing models are primarily tailored for general-purpose tasks like image captioning, generating unstructured natural language outputs that are difficult to integrate into downstream engineering pipelines. Furthermore, without explicit schema-level constraints or domain grounding, current multimodal systems struggle to provide the structured, fine-grained state recognition necessary for actionable real-world applications~\cite{wen2023road}.
        
        To systematically bridge this gap, this paper unleashes the capabilities of large vision–language models in intelligent perception of urban roadside scenarios and infrastructure. Specifically, we unleash the potential of representative models (e.g., Grounding DINO and Qwen-VL) for roadside-related detection, attribute inference, and multimodal reasoning through fine-tuning and schema-guided structured representation. In addition, we incorporate dual-modality knowledge integration via a Retrieval-Augmented Generation (RAG) mechanism that combines textual and visual retrieval. The main contributions of this work are summarized as follows:
        
        \begin{itemize}

        \item \textbf{Enhancing perception and reasoning via targeted fine-tuning strategies:} 
        
        Instead of relying on off-the-shelf models, we unleash the potential of vision-language models (VLMs) through targeted fine-tuning strategies. Specifically, we implement open-vocabulary fine-tuning on Grounding DINO for robust roadside infrastructure detection, and LoRA-based fine-tuning on Qwen-VL for attribute-level reasoning. This design ensures precise object localization and fine-grained interpretation of diverse roadside assets.
        
        \item \textbf{Dual-modality retrieval-augmented generation:} 
        We incorporate professional domain knowledge through a dual-modality Retrieval-Augmented Generation (RAG) mechanism. The proposed approach integrates professional textual knowledge (e.g., GB~5768.2--2022 and related standards) with visual exemplars retrieved from attribute-annotated image repositories, enabling knowledge-grounded and structured perception for roadside asset analysis.
        
        \item \textbf{Urban roadside dataset and comprehensive evaluation:} 
        We construct a large-scale urban roadside dataset with detailed object and attribute-level annotations, covering multiple categories of infrastructure (e.g., traffic signs, lights, bollards, hydrants, and cameras) and their diverse attribute states. Extensive experiments conducted on datasets collected from multiple cities demonstrate the effectiveness and robustness of the proposed framework in fine-grained attribute recognition and structured reasoning tasks.
        
        \end{itemize}

        \section{Related work}

            Intelligent roadside perception fundamentally relies on two core capabilities: the detection of diverse assets and the interpretation of their fine-grained states. First, Sec. \ref{R2.1} traces the shift from rigid, closed-set detectors to flexible open-vocabulary architectures. Second, Sec. \ref{R2.2} examines methodologies for deciphering complex facility conditions, highlighting the recent transition toward semantic reasoning powered by Vision-Language Models.

            \subsection{Infrastructure Open-set Object Detection}\label{R2.1}
                Object detection in urban roadside scenes has long been a cornerstone of ITS perception. Early efforts relied on classical methods such as template matching and feature-based classifiers. With the advent of deep learning, Convolutional Neural Networks (CNNs) and one-stage detectors (e.g., YOLO~\cite{redmon2016you}, SSD~\cite{liu2016ssd}) became the standard, enabling robust detection in complex scenes~\cite{LIU2025106377, ZHOU202263}. While these methods achieve high accuracy on established benchmarks~\cite{HAN2024500}, they are inherently constrained by a closed-set paradigm. Relying on fixed categories defined during training, they lack the flexibility to adapt to dynamic urban environments where novel infrastructure designs frequently emerge.
                
                To overcome the limitations of fixed vocabularies, Open-Vocabulary Detection (OVD) has been developed to recognize objects via arbitrary text queries. Representative approaches, such as GLIP~\cite{li2022grounded} and Grounding DINO~\cite{liu2023improvedllava, zhang2022dino}, reformulate detection as a phrase grounding task, leveraging pre-trained image-text embeddings (e.g., CLIP~\cite{radford2021learning}) to achieve zero-shot generalization. More recently, architectures like YOLO-World~\cite{cheng2024yoloworld} have pushed the boundaries of efficiency in open-set scenarios. However, despite their success on general-purpose benchmarks like COCO, these models are not specifically tailored for the complexities of ITS. Their direct applicability to structured roadside infrastructure remains underexplored, as they often lack the domain-specific granularity required for real-world maintenance, highlighting a critical need for domain-adaptive OVD solutions.
            
            \subsection{Infrastructure Attribute State Recognition}\label{R2.2}
                While category-level detection provides coarse semantic labels, roadside management often requires more fine-grained information regarding the attributes and physical states of infrastructure. For instance, determining whether a traffic light is illuminated or has a countdown timer~\cite{behrendt2017deep}, whether a public trash bin is overflowing~\cite{adedeji2020intelligent}, or whether a traffic sign or bollard is damaged~\cite{tabernik2019deep, houben2019detection} has direct implications for safety and maintenance. In roadside scenes, these requirements are specific and diverse. A traffic light is not only categorized by its type but also by its operational phases, while traffic signs must be assessed for surface damage, fading, or occlusion, each conveying critical operational information.
                
                Existing research on such attribute-level recognition remains sparse and disjointed. Current methodologies primarily focus on isolated tasks, such as specific traffic light state recognition~\cite{electronics13030615} or road surface damage detection~\cite{10.1093/tse/tdac026, Aygun2024Building}, lacking a unified schema that covers multiple infrastructure categories. Furthermore, traditional attribute recognition methods are often limited by closed vocabularies and rigid label definitions, which fail to capture the evolving nature of urban environments.
                
                The emergence of Vision–Language Models (VLMs) offers a promising direction to address these limitations by aligning visual features with semantic understanding~\cite{bordes2024introduction}. The field has advanced rapidly from contrastive models like CLIP~\cite{radford2021learning} to instruction-tuned architectures, such as GPT-4V~\cite{achiam2023gpt4}, LLaVA~\cite{liu2023visual}, and Qwen-VL~\cite{bai2023qwen}. Recent works like TrafficGPT~\cite{zhang2023trafficgpt} have begun to explore their application within the roadside infrastructure domain, utilizing their capabilities in compositional understanding and Visual Question Answering (VQA) to interpret complex scenes.
                
                However, bridging the gap between general-purpose multimodal agents and ITS requirements remains challenging. First, generic VLMs often lack grounding in domain-specific standards, leading to hallucinations or failures in distinguishing functionally similar facilities~\cite{liu2024hallucination}. Second, while VLMs excel at generating descriptive narratives, they predominantly produce unstructured, free-form natural language. Such outputs, although informative, lack the structured, schema-conformant format necessary for direct integration into automated asset monitoring workflows~\cite{Maaz2024Video, chen2024map}.

        \section{Methodology}

            

\begin{figure*}[!t]
                \centering
                \includegraphics[width=\textwidth]{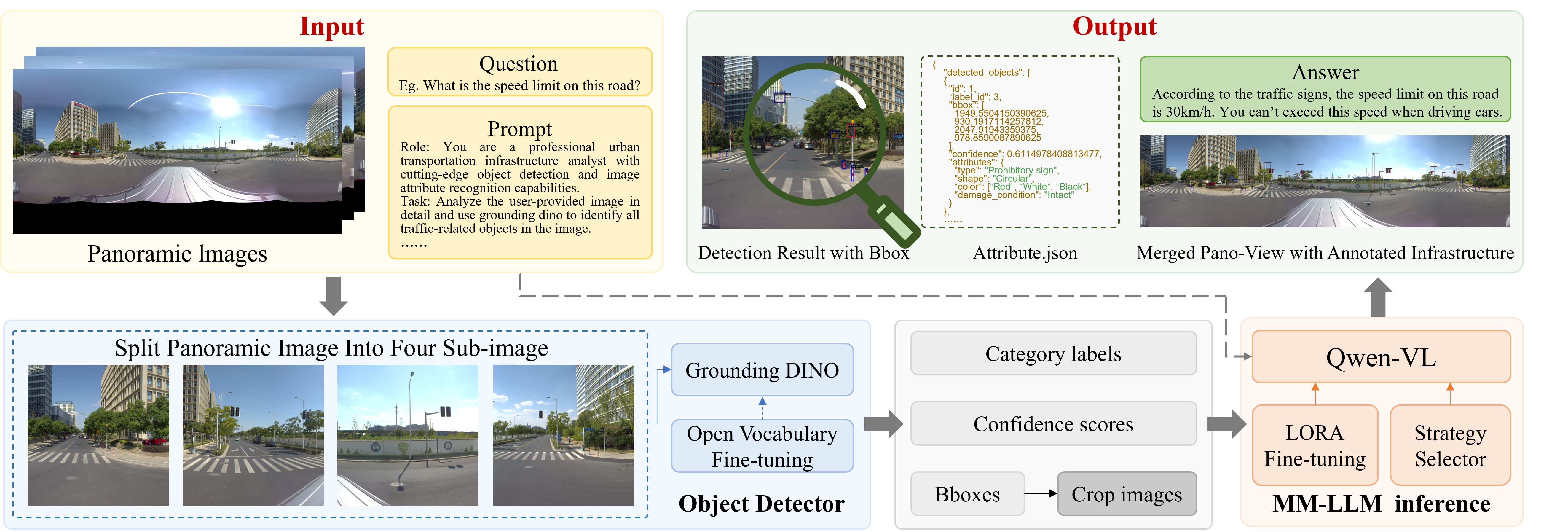} 
                \caption{Workflow of the proposed attribute- and state-aware open-vocabulary vision-language framework.}
                \label{fig_workflow}
                \end{figure*}

            To bridge the gap between general-purpose visual recognition and the rigorous demands of infrastructure maintenance, we present a unified framework that transforms pre-trained Vision-Language Models (VLMs) into domain-specialized agents. As illustrated in Fig. \ref{fig_workflow}, our approach operates as a coarse-to-fine pipeline, integrating precise localization with deep semantic reasoning. The framework consists of three synergistic components: (1) Open-Vocabulary Detection: We employ a fine-tuned Grounding DINO to accurately localize diverse and long-tail roadside assets with high data efficiency; (2) Dual-Modality Knowledge Injection: To mitigate domain gaps, a Retrieval-Augmented Generation (RAG) mechanism dynamically retrieves authoritative industry standards and visual exemplars as contextual prompts; and (3) Attribute-Aware Reasoning: A LoRA-adapted Qwen-VL synthesizes the visual features and retrieved knowledge to infer fine-grained attributes. The following subsections detail the multimodal framework design (Sec. 3.1), the object detection strategy (Sec. 3.2), and the knowledge-enhanced attribute reasoning process (Sec. 3.3).
  
            \subsection{Multimodal Detection-to-Attribute Framework}
                
                The framework integrates the fine-tuned Grounding DINO and Qwen-VL models into a unified multimodal perception and dialogue system. As shown in Fig. \ref{fig_workflow}, the end-to-end process spans open-vocabulary detection to interactive reasoning.
                
                Starting with panoramic images of urban road scenes, the images are divided into four perspective sub-images to improve detection granularity. The Grounding DINO model, fine-tuned for open-vocabulary detection, extracts category labels, confidence scores, and bounding boxes. These boxes are then cropped and processed by the fine-tuned Qwen-VL model, trained with LoRA on the attribute annotation dataset from Section 4.1. Guided by specific prompts, Qwen-VL performs attribute reasoning and generates structured descriptions.
                
\begin{figure}[!h]
                \centering{
                \includegraphics[width=0.48\textwidth]{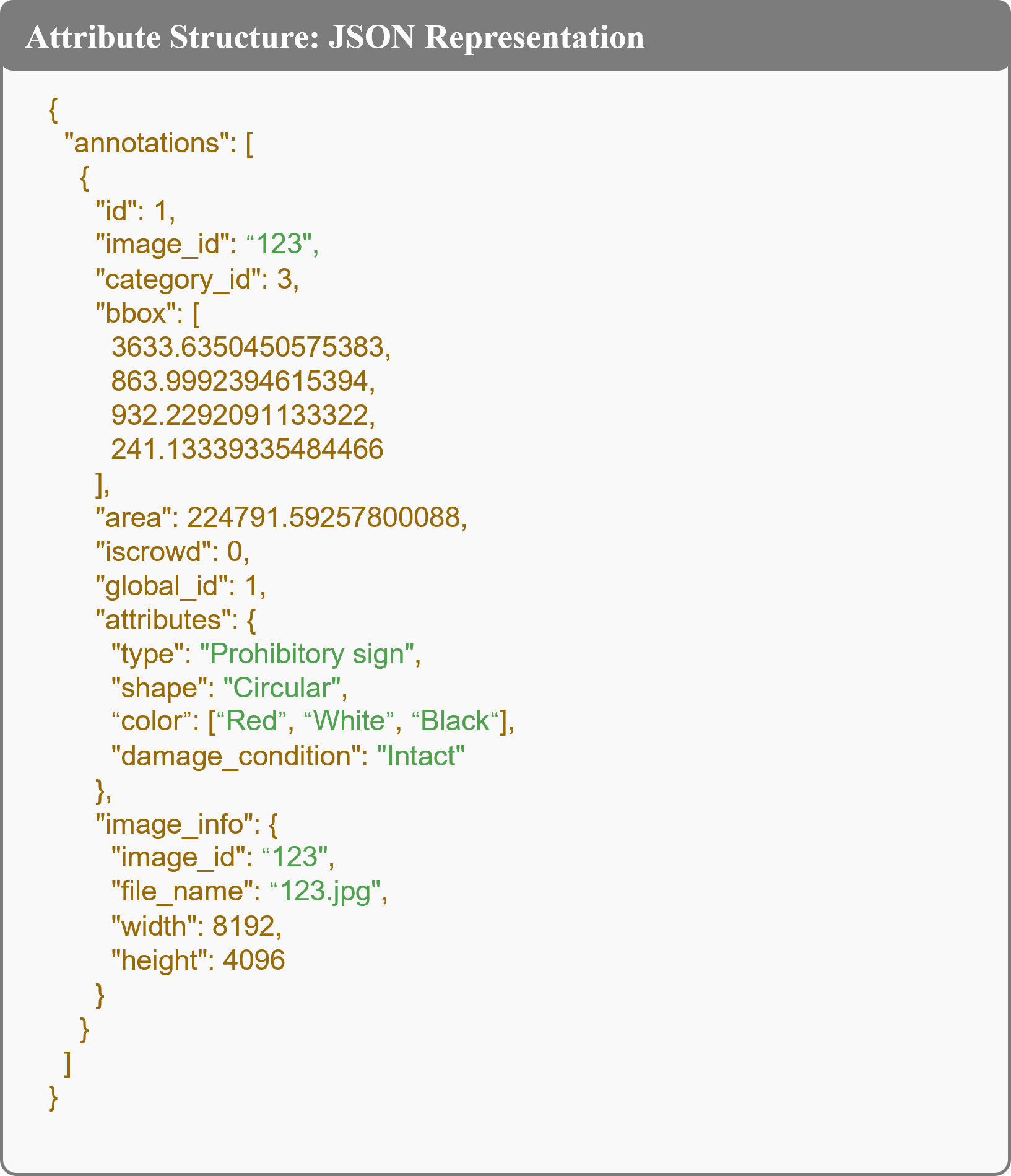}}%
                \caption{Example of structured JSON representation for roadside object attributes.}
                \label{fig_json}
                \end{figure}

                To generate standardized, machine-readable outputs from this vision–language pipeline, we employ a structured JSON representation for roadside scene analysis. Unlike unstructured captions, the JSON format enforces a consistent schema across categories and attributes, ensuring interoperability and supporting large-scale quantitative evaluation. As illustrated in Fig.~\ref{fig_json}, the final JSON output includes each object’s category, bounding box, and attribute–confidence pairs. This unified structure supports both single-image inference and large-scale dataset processing, enabling seamless integration with infrastructure perception and management systems.
                
                The transformation from detection to JSON involves two stages: (\emph{i}) extracting bounding boxes and class labels from Grounding DINO as object instances, and (\emph{ii}) feeding cropped regions with contextual prompts into Qwen-VL to infer fine-grained attributes according to the predefined schema. For simple attributes (e.g., object posture or light state), rule-based cues are embedded within the prompts to ensure stable inference. This two-stage process bridges perception and structured reasoning, allowing the system to output both interpretable and machine-readable results.
                
                Additionally, the fine-tuned Qwen-VL enables multimodal dialogue, allowing users to query specific attributes, states, or conditions of roadside infrastructure directly from the images. The system integrates detection results, structured annotations, and panoramic visualizations into an interactive framework for intelligent urban roadside management. This unified system links visual grounding, attribute reasoning, and human–machine interaction, supporting both automated structured perception and interactive querying.

             \subsection{Roadside Infrastructure Object Detection}
                
\begin{figure}[!h]
            \centering{
            \includegraphics[width=0.50\textwidth]{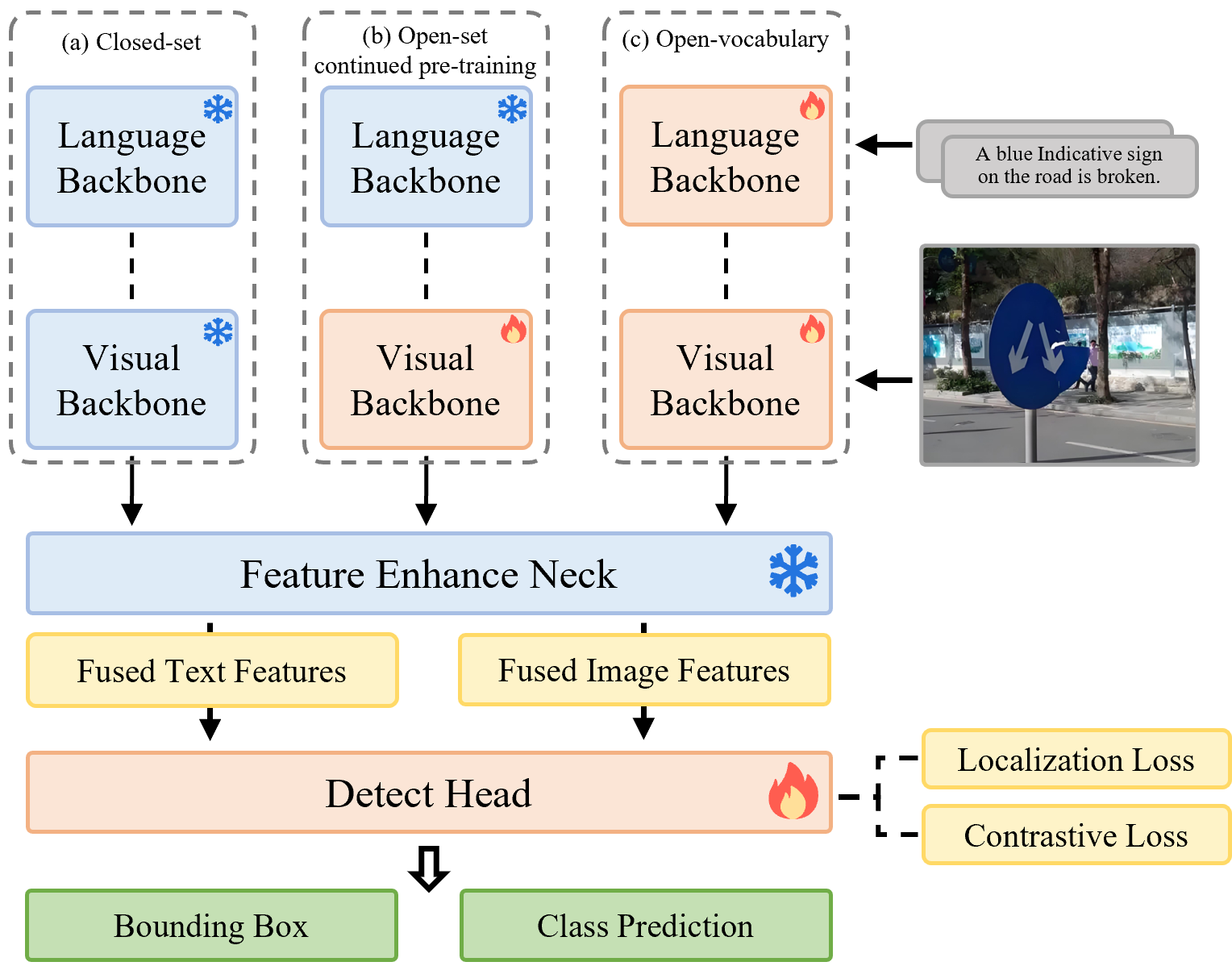}}%
            \caption {Model architectures of the three fine-tuning strategies.}
            \label {fig_FT}
            \end{figure}

                To tailor vision-language models (VLMs) such as Grounding-DINO to the specific requirements of urban roadside scene understanding, we design and evaluate three fine-tuning strategies with different levels of semantic flexibility: closed-set fine-tuning, open-set continued pre-training, and open-vocabulary fine-tuning (OVF). Each strategy employs a distinct model architecture configuration and parameter update policy, reflecting the trade-off between specialization and generalization.The model architecture diagrams of these three fine-tuning strategies are illustrated in Fig. \ref{fig_FT}.

                \textbf{Closed-set fine-tuning:} Closed-set fine-tuning follows the conventional paradigm in which the model is trained only on categories available in the annotated dataset. The detection head is aligned with the target classes, while the visual backbone and language encoder remain frozen to retain pre-trained semantics. This achieves high accuracy for known roadside objects such as traffic signs or traffic lights but fails to recognize novel or evolving facilities.

                \textbf{Open-set continued pre-training fine-tuning:} To improve adaptability while preserving learned knowledge, open-set continued pre-training selectively updates the visual backbone, keeping the language encoder fixed. This allows gradual refinement of visual features and adaptation to new environments while maintaining semantic grounding. Negative sampling is employed to distinguish roadside-relevant regions from background clutter, mitigating hallucination. As a result, this strategy enhances robustness and domain generalization in complex urban scenes.
                
                \textbf{Open-vocabulary fine-tuning:} Building on this, open-vocabulary fine-tuning (OVF) serves as the core of our approach. Rather than relying on fixed class labels, OVF conditions detection on natural language prompts, enabling the model to link unseen visual patterns with descriptive queries. The dataset is divided into Base classes (e.g., vehicles, pedestrians) and Novel classes (e.g., traffic lights, bollards). The model is trained only on Novel classes and evaluated jointly on both, testing its generalization capability. OVF jointly updates the backbone and language encoder with controlled learning rates to mitigate catastrophic forgetting. This design maintains open-domain knowledge while enhancing generalization to unseen roadside categories via descriptive prompts.

                In summary, while closed-set fine-tuning provides strong accuracy for fixed categories and open-set pre-training improves robustness, open-vocabulary fine-tuning uniquely maintains prior knowledge while enabling scalable detection of unseen objects—making it the most effective strategy for real-world intelligent roadside systems where infrastructure evolves continuously.

        \subsection{Roadside Infrastructure Attribute State Recognition}

\begin{figure*}[!h]
            \centering{
            \includegraphics[width=0.9\textwidth]{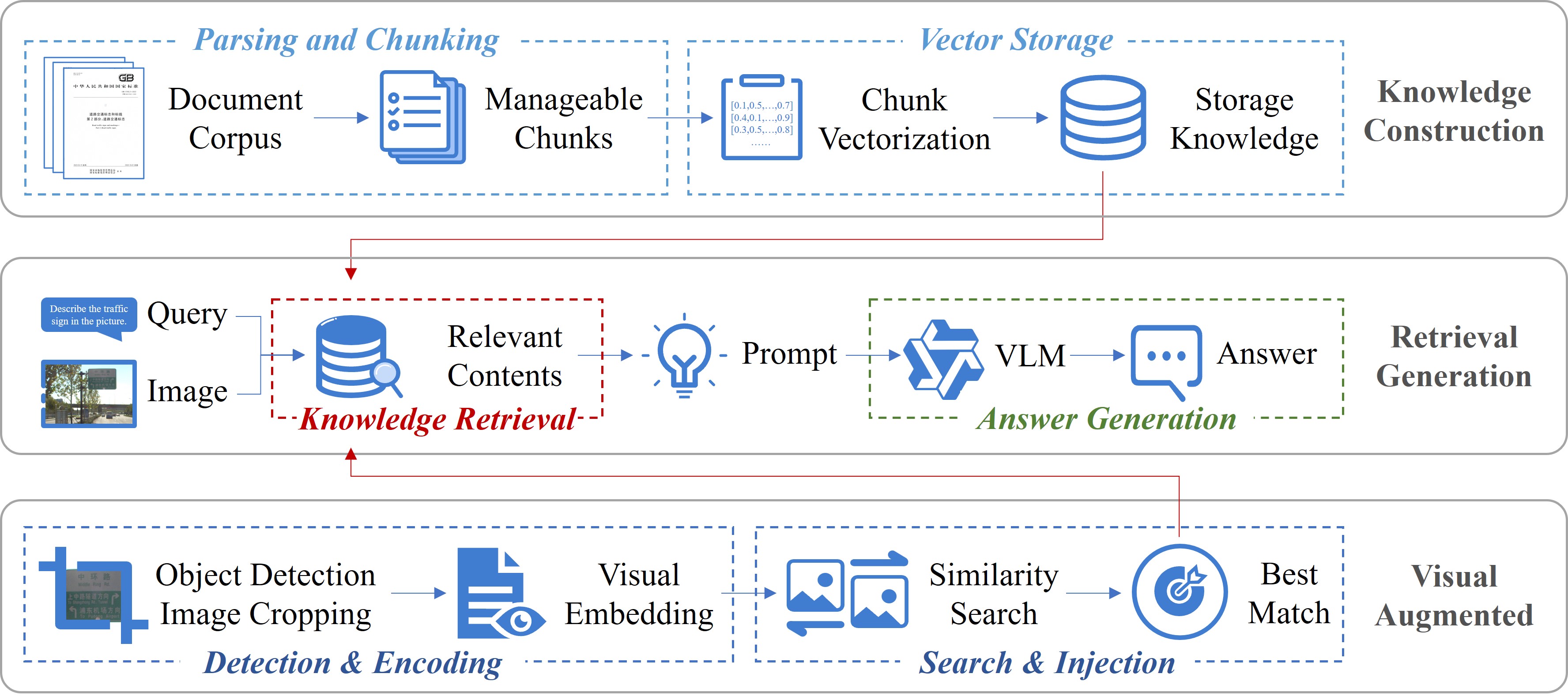}}%
            \caption{Overview of the dual Retrieval-Augmented Generation (RAG) framework, integrating Textual RAG for knowledge-grounded reasoning using domain standards, and Visual RAG for exemplar-based attribute retrieval from annotated image databases.}
            \label{fig_rag}
            \end{figure*}

           To enable precise fine-grained attribute recognition for roadside infrastructure, we introduce a specialized module that adapts Vision-Language Models into domain-expert agents. As shown in Fig.~\ref{fig_rag}, it integrates attribute-guided fine-tuning with dual-modality Retrieval-Augmented Generation (RAG). The module comprises two components: (1) Attribute-Guided Fine-Tuning: LoRA adaptation of Qwen-VL aligned to structured attribute schemas via visual instruction tuning; and (2) Dual-Modality RAG: Retrieval of textual standards and visual exemplars to ground attribute inference in domain knowledge. The following subsections describe the fine-tuning strategy (Sec. 3.3.1) and the RAG-enhanced reasoning (Sec. 3.3.2).

            \subsubsection{Attribute-Guided Fine-tuning for VLMs}
    
                Although pre-trained vision–language models possess strong general perception capabilities, they often lack the domain-specific granularity required to accurately interpret complex roadside assets and struggle to strictly adhere to predefined attribute schemas~\cite{bai2023qwen_tech}. To address these limitations and adapt the general-purpose model into a domain specialist, we implement an attribute-guided fine-tuning strategy.
                
                Building upon the structured attribute annotations, we fine-tune Qwen-VL to improve attribute reasoning and domain adaptation performance. The core objective is to enable the model to parse roadside infrastructure states with high precision while maintaining rigorous output formatting. To achieve this, we formulate the attribute recognition task as a supervised visual instruction tuning problem~\cite{liu2023improvedllava}.

                The annotated data described in Section 4.1 is transformed into instruction-following pairs, structured as $\langle \text{Image}, \text{Instruction}, \text{Output} \rangle$. For each training instance, the input consists of the cropped object image and a prompt template (e.g., \emph{``Identify the attributes of the roadside object in the image following the standard schema.''}), while the target output is the ground-truth JSON sequence. This format forces the model to learn the correspondence between visual features and the specific attribute schema defined for urban infrastructure.
                
                Fine-tuning is conducted using parameter-efficient Low-Rank Adaptation (LoRA)~\cite{hu2021lora}, which minimizes computational overhead while preserving the model's generalization capability. Instead of updating the entire parameter space of the Large Language Model (LLM), we freeze the pre-trained vision encoder and the majority of the LLM backbone. LoRA adapters are injected specifically into the projection layers (e.g., query, key, value projections) of the attention mechanisms within the transformer blocks.
                
                Mathematically, for a pre-trained weight matrix \( W \), LoRA introduces a low-rank decomposition structure:
                \begin{equation}
                W' = W + \Delta W = W + A B
                \end{equation}
                where \( A \) and \( B \) are low-rank matrices, and \( \Delta W \) represents the update learned during fine-tuning. This structure allows the model to adapt efficiently to domain-specific knowledge—such as distinguishing specific sign types or damage conditions—without suffering from catastrophic forgetting of its foundational capabilities.
                
                During training, the model is supervised using the structured JSON sequences. The optimization objective is to maximize the probability of generating the correct attribute tokens and structure markers given the visual input and textual instructions. This focused supervision enables the model to transition from generating free-form descriptions to producing rigorous, schema-conformant outputs suitable for automated engineering downstream tasks.
                
                After fine-tuning, Qwen-VL exhibits enhanced consistency between perception and reasoning modules. When integrated with the fine-tuned Grounding DINO, the system demonstrates improved attribute recognition accuracy, reduced hallucination, and stronger robustness under diverse urban conditions.

            \subsubsection{Domain Knowledge Imbedding via RAG}
    
                To compensate for the dual gap in domain-specific knowledge and exemplar-based visual experience that limits reliable attribute inference, we incorporate a dual-modality Retrieval-Augmented Generation (RAG) mechanism that integrates both professional textual knowledge and visual exemplars into the multimodal reasoning process. RAG~\cite{lewis2020rag} combines external information retrieval with generative modeling, allowing large vision–language models to access structured knowledge and historical visual experience during inference.
                
                As illustrated in Fig.~\ref{fig_rag}, the proposed framework includes two complementary RAG branches: (1) a \emph{textual RAG} for integrating domain-specific documents, and (2) a \emph{visual RAG} for leveraging image-based retrieval to enhance object-level attribute reasoning.
                
                \textbf{Textual RAG for Knowledge-Grounded Reasoning:}  
                In the textual RAG branch, a \emph{domain knowledge base} is constructed by parsing, segmenting, and encoding authoritative references such as roadside regulations, signage standards, and infrastructure manuals. National and regional standards (e.g., GB~5768.2–2022 \emph{Road Traffic Signs} and related specifications) are transformed into structured vector embeddings and stored in a local semantic database for retrieval~\cite{pan2024vectorrag}. 
                
                Given a user query $q_{\text{text}}$ related to attribute inference (e.g., \emph{``Is this a warning sign?''}), the system performs semantic matching to retrieve the top-$k$ most relevant text fragments $\{d_1, d_2, \ldots, d_k\}$ from the knowledge base $\mathcal{D}$. The retrieval process is formulated as:
                \begin{equation}
                \{d_1, \ldots, d_k\} = \text{Top-}k\left(\left\{\text{sim}(\mathbf{e}_q, \mathbf{e}_{d_i}) \mid d_i \in \mathcal{D}\right\}\right),
                \end{equation}
                where $\mathbf{e}_q$ and $\mathbf{e}_{d_i}$ denote the text embeddings of the query and document fragment $d_i$, respectively, and $\text{sim}(\cdot, \cdot)$ represents the cosine similarity:
                \begin{equation}
                \text{sim}(\mathbf{e}_q, \mathbf{e}_{d_i}) = \frac{\mathbf{e}_q \cdot \mathbf{e}_{d_i}}{\|\mathbf{e}_q\| \|\mathbf{e}_{d_i}\|}.
                \end{equation}
                
                These retrieved definitions and rules are dynamically appended to the model's contextual input, enabling the multimodal model (Qwen-VL) to perform \emph{knowledge-grounded reasoning}. This process ensures that predictions adhere to formal domain conventions, such as recognizing that ``warning signs'' typically adopt \emph{yellow triangular} configurations, as defined in GB~5768.2–2022 and related standards.

                \textbf{Visual RAG for Object-Level Attribute Retrieval:}  
                In addition to textual grounding, we design a visual retrieval-augmented generation pipeline that enhances object-level attribute recognition through exemplar-based reasoning~\cite{zhou2024visual}. For each detected roadside object, the system first encodes its cropped image $I_{\text{crop}}$ into a visual embedding $\mathbf{v}_{\text{query}}$ using a pre-trained CLIP~\cite{radford2021learning} vision encoder:
                \begin{equation}
                \mathbf{v}_{\text{query}} = f_{\text{CLIP}}(I_{\text{crop}}),
                \end{equation}
                where $f_{\text{CLIP}}(\cdot)$ represents the CLIP image encoder that maps the cropped image into a normalized feature space.
                
                Subsequently, a similarity search is performed within an attribute-annotated image database $\mathcal{V} = \{(I_i, a_i)\}_{i=1}^{N}$, where $I_i$ denotes the $i$-th reference image and $a_i$ represents its corresponding attribute annotations. The retrieval identifies the top-$m$ most visually similar samples based on cosine similarity:
                \begin{equation}
                \{(I_1^*, a_1^*), \ldots, (I_m^*, a_m^*)\} = \text{Top-}m\left(\left\{\text{sim}(\mathbf{v}_{\text{query}}, \mathbf{v}_i) \mid (I_i, a_i) \in \mathcal{V}\right\}\right),
                \end{equation}
                where $\mathbf{v}_i = f_{\text{CLIP}}(I_i)$ denotes the pre-computed embedding of reference image $I_i$, and the cosine similarity is computed as:
                \begin{equation}
                \text{sim}(\mathbf{v}_{\text{query}}, \mathbf{v}_i) = \frac{\mathbf{v}_{\text{query}} \cdot \mathbf{v}_i}{\|\mathbf{v}_{\text{query}}\| \|\mathbf{v}_i\|}.
                \end{equation}
                
                The retrieved attribute annotations $\{a_1^*, \ldots, a_m^*\}$ are then injected as contextual inputs into the multimodal model. This process enables Qwen-VL to reference prior visual–semantic correspondences, leveraging historical annotation experience to produce more accurate and consistent attribute predictions. For example, when identifying a "spherical bollard with reflective strips," the retrieved exemplars provide prior knowledge about surface reflectivity or damage patterns, improving fine-grained attribute reasoning.
                
                \textbf{Unified Knowledge-Augmented Generation:}  
                
\begin{figure}[!h]
                \centering
                \includegraphics[width=0.45\textwidth]{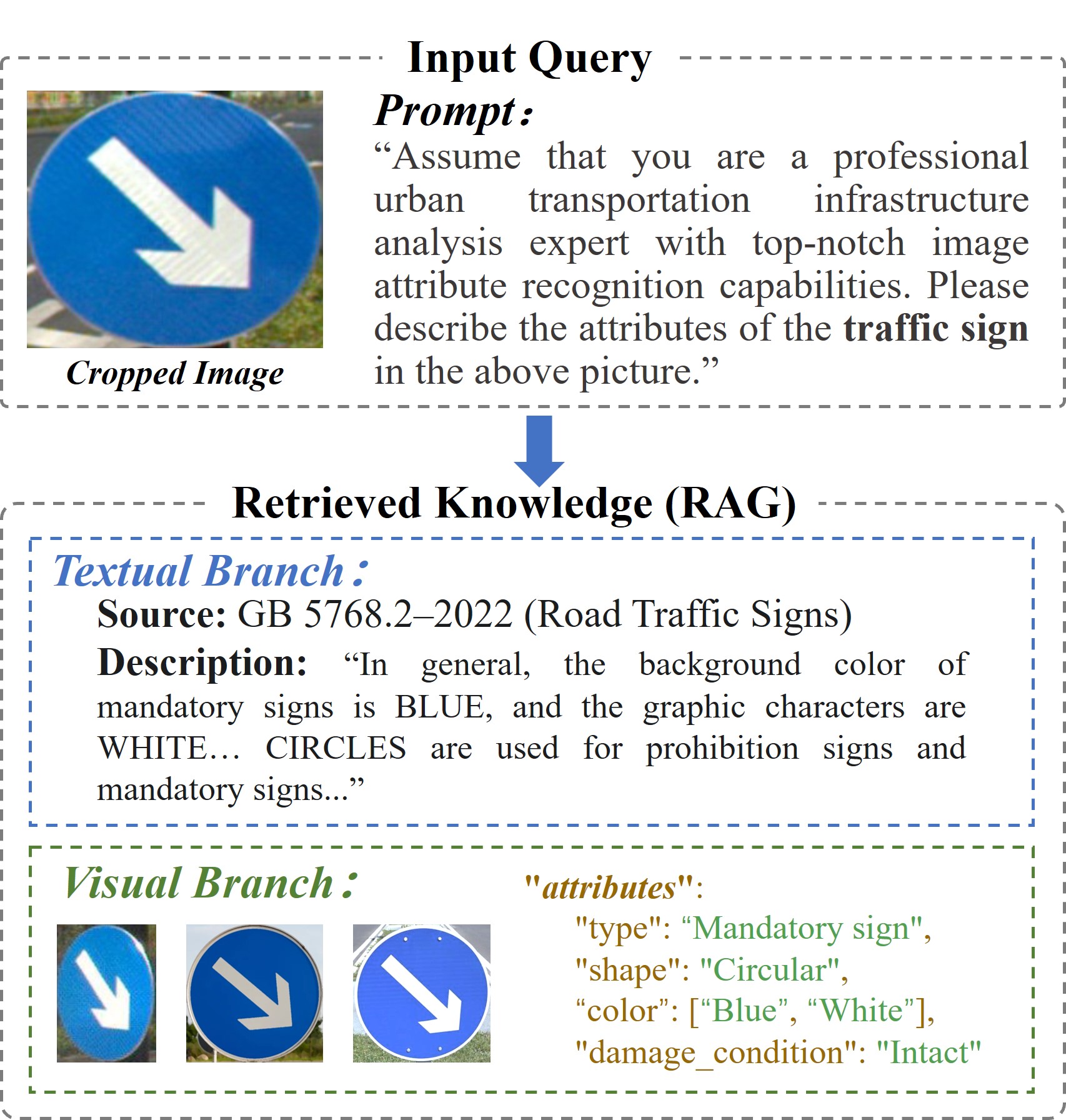}
                \caption{A case study of the RAG-enhanced inference process. The system retrieves the ``Mandatory sign'' definition from GB 5768.2–2022 and visually similar exemplars to guide the generation of structured attribute outputs for a blue circular traffic sign.}
                \label{fig_rag_case}
                \end{figure}

                By integrating both textual and visual retrieval, the system transitions from perceptual pattern recognition to knowledge-grounded, context-enriched reasoning. A concrete inference example is presented in Fig.~\ref{fig_rag_case}, where the retrieval of the standard definition (e.g., circular shape, blue background) and visually similar exemplars effectively guides the model to generate precise, schema-conformant attributes for a mandatory traffic sign. The final input to the vision-language model combines the original image, the retrieved textual knowledge $\{d_1, \ldots, d_k\}$, and the visual exemplar annotations $\{a_1^*, \ldots, a_m^*\}$, formulated as:
                \begin{equation}
                \text{Output} = \text{VLM}(I_{\text{crop}}, I_{\text{original}}, \{d_1, \ldots, d_k\}, \{a_1^*, \ldots, a_m^*\}),
                \end{equation}
                where $\text{VLM}(\cdot)$ denotes the vision-language model that performs multimodal reasoning over the augmented context.
                
                The textual RAG provides formal semantic constraints derived from standards, while the visual RAG supplies perceptual analogs drawn from prior samples. Together, they enhance interpretability, reduce hallucination, and ensure that generated attributes are both visually grounded and semantically compliant with professional definitions. Moreover, new documents or annotated samples can be continuously added to the knowledge base without retraining, ensuring scalability and adaptability for evolving roadside infrastructures.

        \section{Experiments and Results}

             This section presents a comprehensive evaluation of the proposed framework. We begin by introducing the annotated dataset and evaluation metrics, followed by an assessment of the system's capabilities in open-vocabulary detection and fine-grained attribute recognition. Finally, we conduct extensive comparative experiments against representative baselines to validate the effectiveness of the proposed fine-tuning and retrieval-augmented strategies.
             
            \begin{figure*}[!t]
            \centering
            \includegraphics[width=\textwidth]{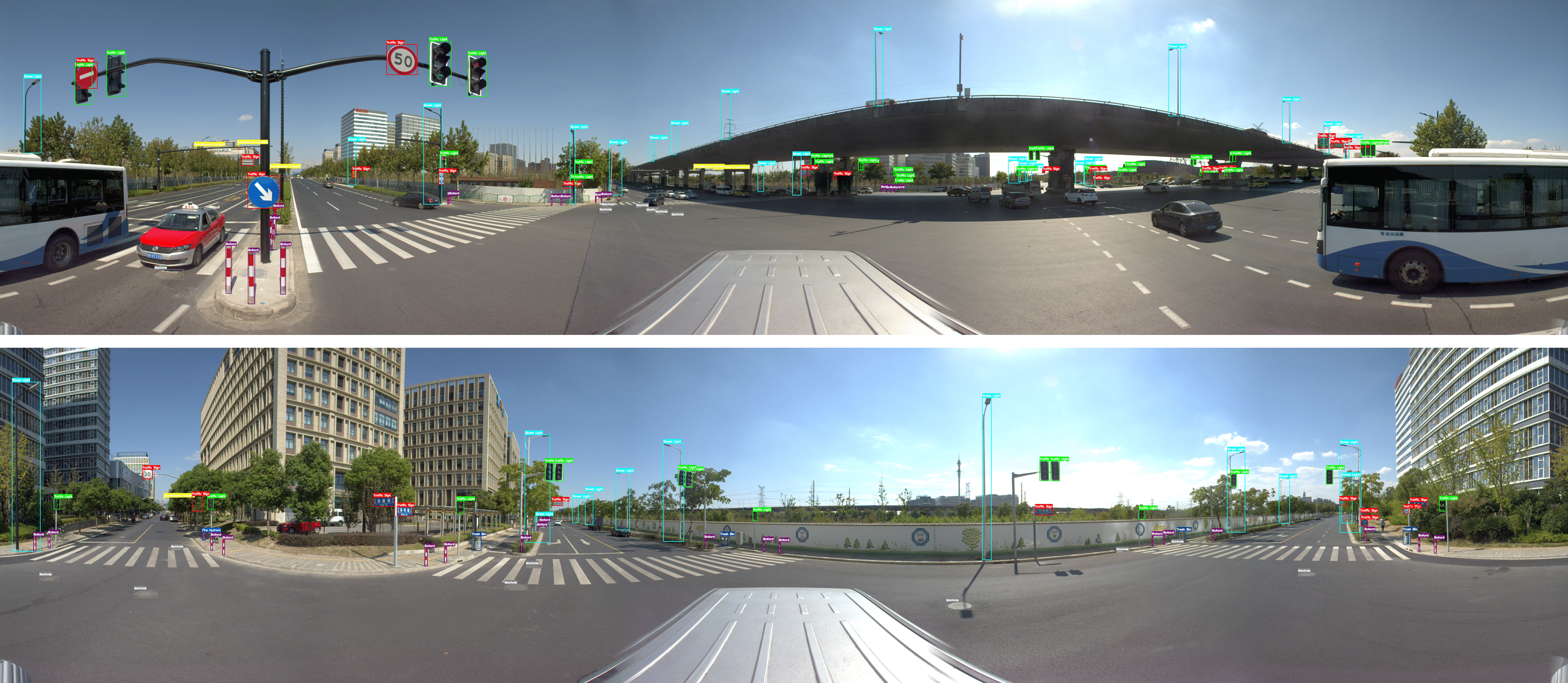}
            \caption{Examples of object detection annotations from the roadside infrastructure dataset in urban roadside scenes.}
            \label{fig_detection_annotations}
            \end{figure*}

            \begin{figure}[t!]
            \centering
            \includegraphics[width=\linewidth]{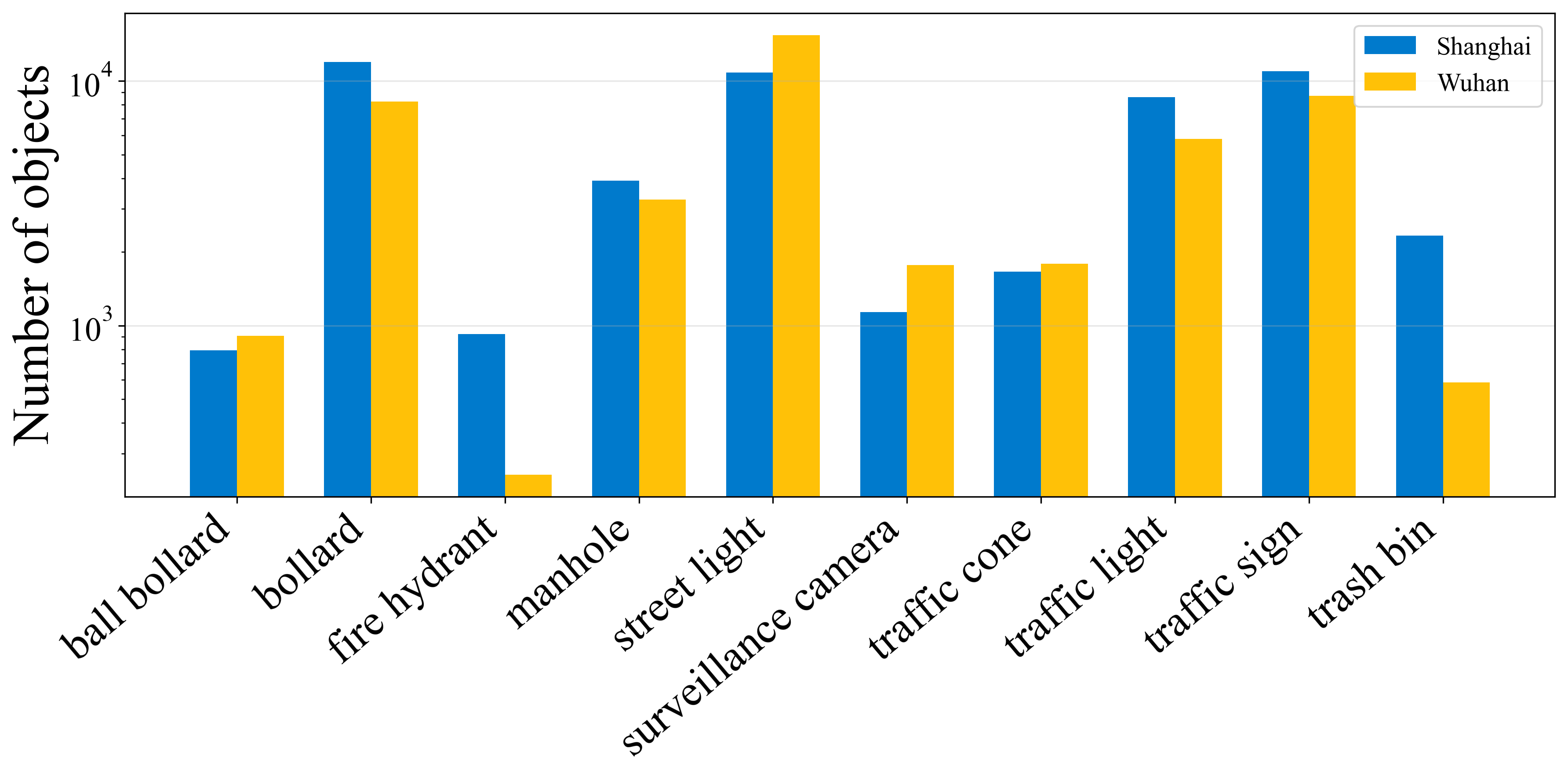}
            \caption{The distribution of annotation numbers among different semantic categories in our dataset. The blue and yellow bars represent the statistics for the Shanghai and Wuhan subsets, respectively.}
            \label{fig_annotation_distribution}
            \end{figure}

                \subsection{Attribute-Annotated Roadside Infrastructure Dataset}
    
                    To facilitate fine-grained perception and multimodal reasoning within complex urban environments, we construct a large-scale \textbf{urban roadside infrastructure dataset} enriched with object-level annotations and fine-grained attribute labels. The dataset serves as a comprehensive benchmark for both perception and attribute reasoning tasks within complex urban scenes, advancing beyond traditional benchmarks that focus primarily on segmentation or coarse detection~\cite{cordts2016cityscapes, yu2020bdd100k}.
                    
                    The dataset is constructed from panoramic roadside images captured by vehicle-mounted sensors across two major Chinese cities: Shanghai and Wuhan. In total, the dataset contains 3,551 high-resolution panoramic images (8192×4096), comprising 1,576 images from Shanghai and 1,975 from Wuhan. For the Shanghai subset, we use 1,316 images for training and 260 images for validation. For the Wuhan subset, the training split includes 1,746 images, while the validation split contains 229 images. This city-specific partitioning supports the subsequent experiments, where each city is evaluated independently as well as under cross-city generalization settings.
                    
\begin{figure*}[!t]
                    \centering
                    \includegraphics[width=\textwidth]{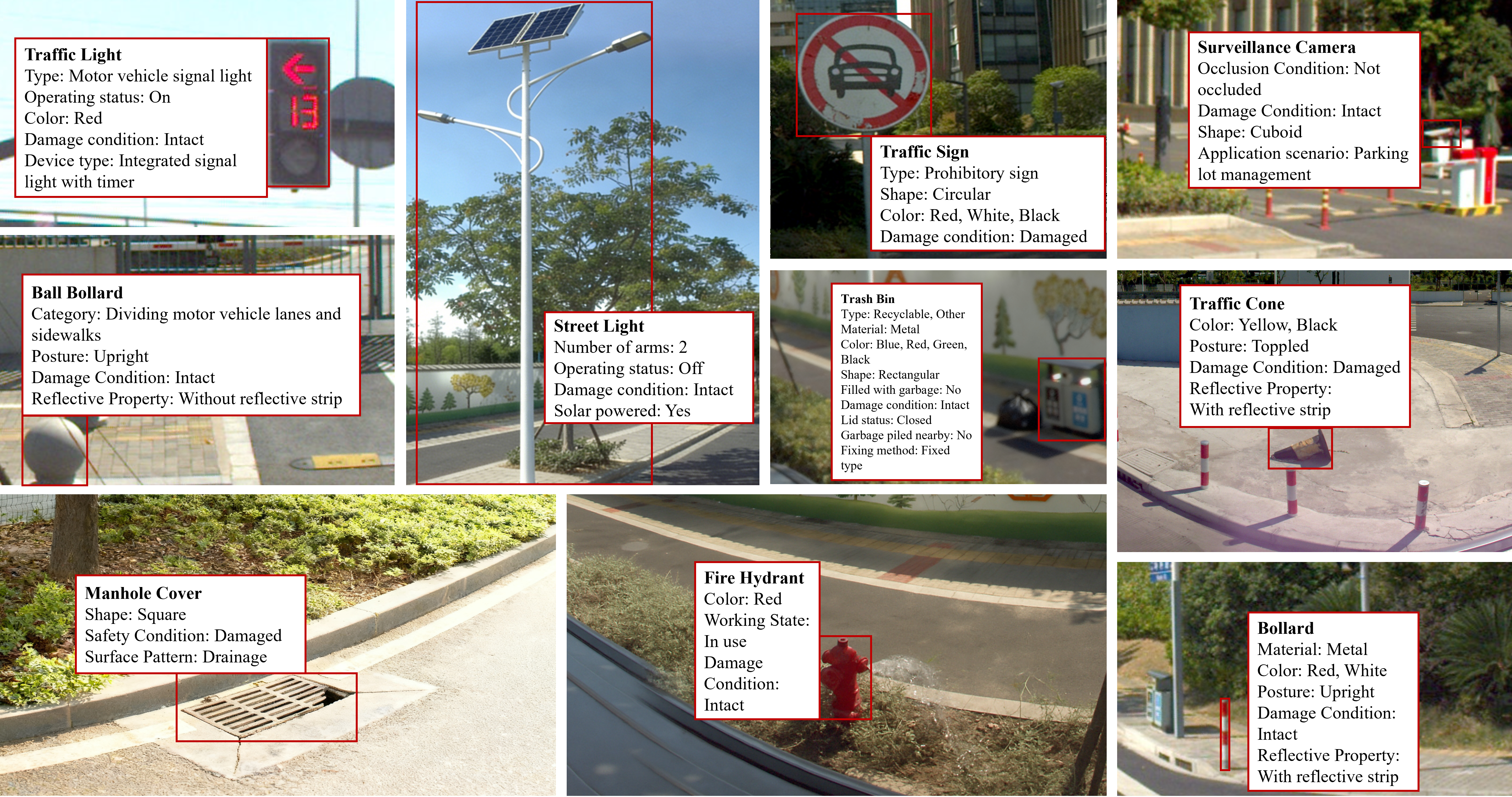}
                    \caption{Examples from the roadside infrastructure dataset, illustrating diverse categories, illumination conditions, and attribute variations.}
                    \label{fig_dataset}
                    \end{figure*}

                    This dataset includes more than 100,000 annotated instances across ten representative infrastructure categories: traffic signs, signal lights, street lights, surveillance cameras, bollards, ball bollards, fire hydrants, trash bins, manhole covers, and traffic cones. The quantitative distribution of annotations across these categories for both Shanghai and Wuhan is illustrated in Fig.~\ref{fig_annotation_distribution}. These annotations provide rich object-level and attribute-level information, supporting a variety of perception and reasoning tasks. Examples of the object detection annotations across these categories are shown in Fig.~\ref{fig_detection_annotations} and Fig.~\ref{fig_dataset}, illustrating the annotated objects in urban roadside scenes.

                    To support fine-grained and interpretable annotation of roadside infrastructure, we construct a structured attribute schema that organizes common urban roadside objects and their associated properties. As detailed in Table~\ref{table:attribute_schema}, the dataset covers ten representative categories, where each category is assigned a set of practical semantic attributes such as shape, color, material, operational status, and physical condition.
        
                    Unlike conventional datasets that provide only object labels, our schema offers clear definitions for each attribute and its possible values. This helps ensure annotation consistency and provides a detailed reference for subsequent model training and evaluation. During annotation, the schema serves as a guideline for describing what information should be recorded for each object instance, enabling the dataset to capture not only object categories but also their states and functional characteristics.
                    
                    By establishing this attribute-centric structure, we bridge the gap between low-level perception and high-level semantic reasoning, thereby improving interpretability and supporting structured interaction in downstream roadside analysis systems.
                    
                    \begin{table}[!h]
                    \centering
                    \caption{Attribute schema for ten categories of urban roadside facilities.}
                    \label{table:attribute_schema}
                    \resizebox{\linewidth}{!}{%
                    \begin{tabular}{p{3.0cm} p{7.0cm}}  
                    \toprule
                    \textbf{Category} & \textbf{Attributes} \\
                    \midrule
                
                    Traffic Light & Type, Working State, Color, Damage Condition, Device Type ... \\
                    \midrule
                
                    Street Light & Number of Arms, Working State, Damage Condition, Solar-Powered ... \\
                    \midrule
                
                    Traffic Sign & Type, Shape, Color, Damage Condition ... \\
                    \midrule
                
                    Bollard & Material, Color, Posture, Damage Condition, Reflective Property ... \\
                    \midrule
                
                    Ball Bollard & Category, Posture, Damage Condition, Reflective Property ... \\
                    \midrule
                
                    Surveillance Camera & Occlusion Condition, Damage Condition, Shape, Application scenario ... \\
                    \midrule
                
                    Manhole Cover & Shape, Safety Condition, Surface Pattern ... \\
                    \midrule
                
                    Trash Bin & Category, Material, Color, Shape, Fullness, Damage Condition, Lid Condition, Nearby Garbage Piles, Fixed Type ... \\
                    \midrule
                
                    Fire Hydrant & Color, Working State, Damage Condition ... \\
                    \midrule
                
                    Traffic Cone & Color, Posture, Damage Condition, Reflective Property ... \\
                    \bottomrule
                    \end{tabular}%
                    }
                \end{table}

            \subsection{Performance of Roadside Infrastructure Detection}
                
\begin{figure}[!h]
                \centering{
                \includegraphics[width=0.48\textwidth]{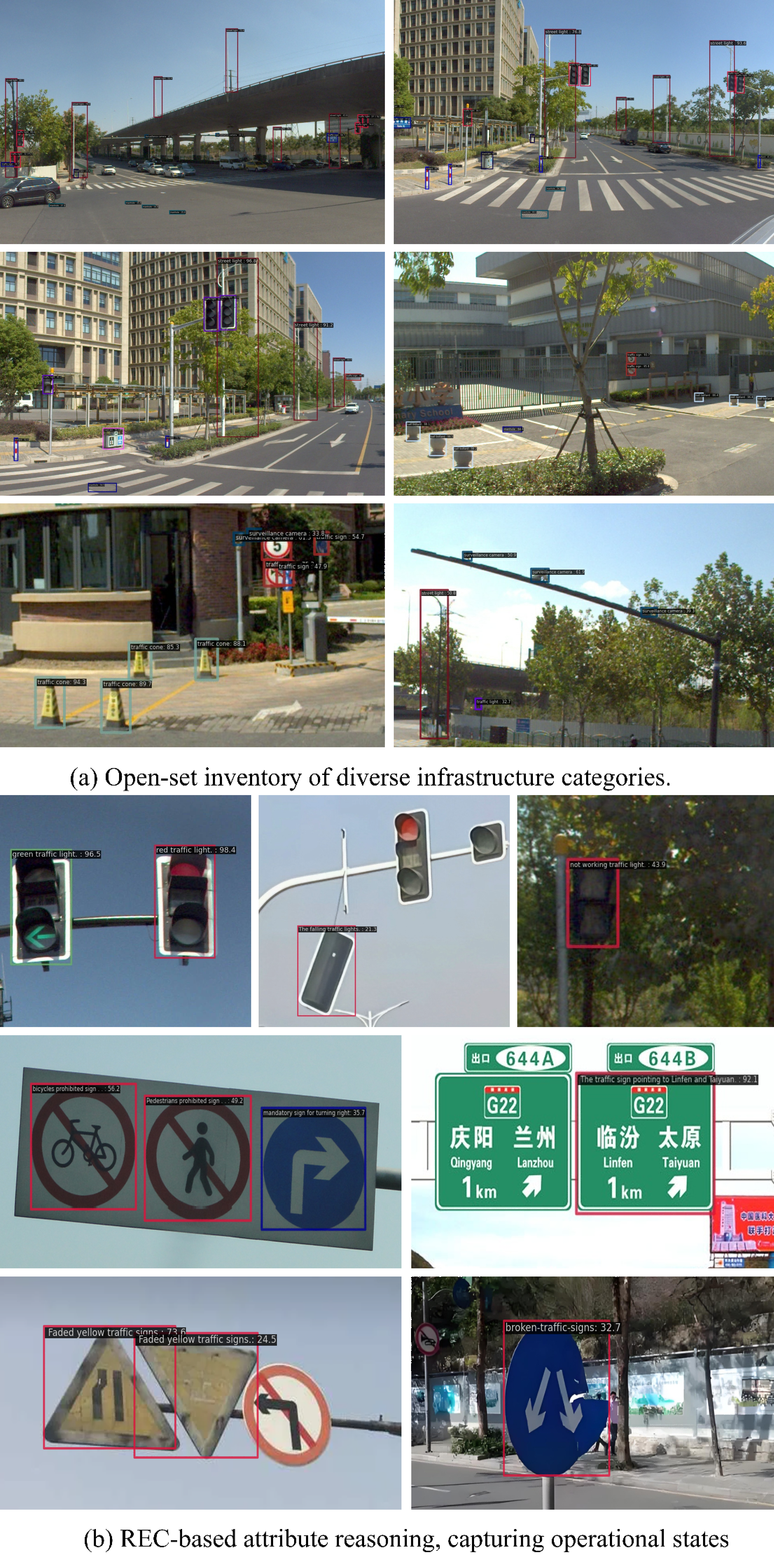}}%
                \caption{Open-vocabulary perception and fine-grained state analysis of roadside infrastructure. (a) Open-set inventory of diverse infrastructure categories. (b) REC-based attribute reasoning, capturing operational states (e.g., signal colors, malfunctions), physical defects (e.g., damage, fading), and semantic interpretations (e.g., sign types and directionality).}
                \label{fig_open_set_inventory}
                \end{figure}

                In the open-set setting, open-vocabulary detectors transcend the limitation of recognizing only classes predefined in a fixed training set. This capability is fundamental for building a comprehensive inventory of urban roadside infrastructure, where object categories can be diverse and evolving. In this experiment, we evaluate the open-vocabulary detection performance in complex roadside scenes. As shown in Fig.~\ref{fig_open_set_inventory}(a), the model effectively leverages language-guided grounding to identify standard roadside-related objects such as traffic signs, signal lights, and bollards within cluttered urban environments. Beyond these common entities, the detector also generalizes to a broader range of facilities, including surveillance cameras, barriers, fire hydrants, trash bins, and manhole covers. This demonstrates the flexibility of language-guided grounding in handling diverse infrastructure categories based on semantic descriptions rather than rigid classes.
                                        
                Furthermore, leveraging its Referring Expression Comprehension (REC) capabilities, the model can identify objects based on specific attributes or conditions, which is crucial for dynamic scene analysis. As illustrated in the detailed views of Fig.~\ref{fig_open_set_inventory}(b), for roadside infrastructure, the model successfully identifies traffic lights and differentiates their operational states such as red, green, and yellow. It also detects abnormal conditions including not-working or falling lights. In terms of traffic sign perception, the model distinguishes prohibitory, warning, and mandatory signs, and further detects defects such as faded or damaged signs. These abilities are important for facility maintenance and roadside regulation management.
                
                While this detection-level attribute awareness provides immediate visual alerts for maintenance, it is primarily unstructured and lacks professional domain knowledge. To achieve standardized and machine-readable asset management, these detection results serve as the pivotal input for the subsequent multimodal reasoning module, which generates the structured JSON outputs detailed in the following sections.

                To strictly quantify the reliability of these detections before they are utilized for downstream reasoning, we adopt standard object detection metrics commonly used in the literature.
                
                For detection accuracy, we first employ Intersection over Union (IoU), which quantifies the spatial overlap between a predicted bounding box and its corresponding ground-truth annotation:
                \begin{equation}
                    \begin{aligned}
                        \text{IoU} &= \frac{Area\ of\ Union}{Area\ of\ Overlap} \\
                    \end{aligned}
                    \label{metric}
                \end{equation}
                
                IoU ranges from 0 to 1, where larger values indicate more accurate localization. In practice, detections are considered correct when IoU exceeds predefined thresholds (e.g., 0.5 or 0.75), reflecting different strictness levels of spatial alignment.
                
                Based on IoU-matched predictions, we further report Precision and Recall to characterize detection reliability and completeness:
                \begin{equation}
                    \begin{aligned}
                        \text{Precision} &= \frac{TP}{TP + FP} \\
                        \text{Recall} &= \frac{TP}{TP + FN} \\
                    \end{aligned}
                    \label{metric}
                \end{equation}
                Precision evaluates the proportion of correct predictions among all detected objects, while Recall measures the proportion of ground-truth objects that are successfully detected. Together, they reflect the trade-off between false alarms and missed detections in real-world roadside scenes.
                
                To provide a unified measure over different recall levels, we adopt Average Precision (AP), which summarizes the precision–recall curve as:
                \begin{equation}
                    \begin{aligned}
                        AP = \int_0^1 P(r)dr,
                    \end{aligned}
                    \label{metric}
                \end{equation}
                where $P(r)$ denotes the precision at recall level $r$. In this work, AP follows the definition and numerical integration protocol established by the COCO benchmark, and detailed computation procedures can be found in the original COCO evaluation paper~\cite{lin2014microsoft}.
                
                Finally, Mean Average Precision (mAP) is reported as the mean AP across all evaluated categories and serves as the primary indicator of overall detection performance. We adopt commonly used variants, including mAP@50 and mAP@75, which correspond to IoU thresholds of 0.5 and 0.75, respectively, as well as mAP@50:95, which averages AP over IoU thresholds from 0.5 to 0.95 with a step size of 0.05. This multi-threshold evaluation provides a comprehensive assessment of detection robustness under varying localization requirements.
                
                Applying these metrics to our collected data, we first analyze the fine-grained performance across different infrastructure types. Table~\ref{tab:per_class_detection} details the per-class detection accuracy on the Shanghai and Wuhan datasets after fine-tuning. It is observed that while most categories achieve satisfactory results, certain classes like surveillance cameras and manholes show relatively lower mAP. This performance gap is primarily due to their smaller physical scale and the challenge of identifying subtle visual features from fixed roadside perspectives.

                \begin{table}[!h]
                \centering
                \caption{Per-class detection performance of roadside infrastructure categories on the Shanghai and Wuhan datasets after fine-tuning.}
                \label{tab:per_class_detection}
                \resizebox{\linewidth}{!}{%
                \begin{tabular}{lcccc}
                \toprule
                \multirow{2}{*}{\textbf{Category}} & \multicolumn{2}{c}{\textbf{Shanghai}} & \multicolumn{2}{c}{\textbf{Wuhan}} \\
                \cmidrule(lr){2-3} \cmidrule(lr){4-5}
                 & \textbf{mAP} & \textbf{mAP@50} & \textbf{mAP} & \textbf{mAP@50} \\
                \midrule
                Traffic light       & 68.3 & 87.7 & 67.2 & 89.7 \\
                Fire hydrant        & 56.0 & 92.7 & 51.4 & 80.9 \\
                Street light        & 61.5 & 79.9 & 64.2 & 81.9 \\
                Traffic sign        & 63.9 & 85.5 & 63.8 & 77.7 \\
                Bollard             & 49.4 & 89.7 & 62.1 & 90.7 \\
                Surveillance camera & 30.2 & 65.3 & 39.7 & 75.9 \\
                Manhole             & 49.5 & 70.2 & 44.2 & 68.7 \\
                Trash bin           & 49.5 & 68.8 & 73.8 & 91.9 \\
                Ball bollard        & 58.7 & 78.7 & 60.0 & 80.8 \\
                Traffic cone        & 49.8 & 84.6 & 62.5 & 87.0 \\
                \midrule
                All                 & 53.2 & 80.3 & 58.9 & 82.5 \\
                \bottomrule
                \end{tabular}%
                }
                \end{table}

                \begin{figure*}[!t]
                \centering{
                \includegraphics[width=\textwidth]{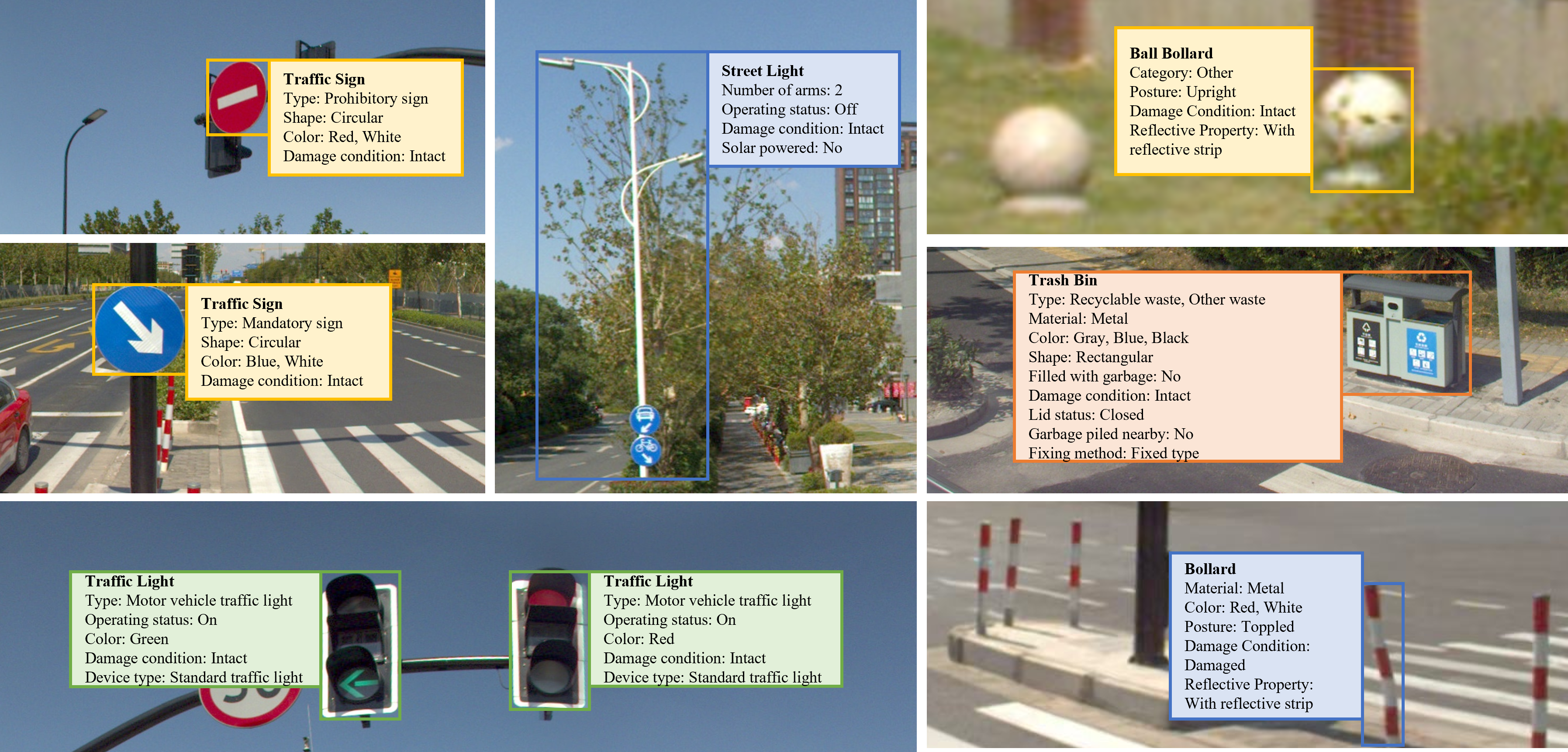}}%
                \caption{Example of Attribute Recognition in Roadside Infrastructure.}
                \label{fig_attribute}
                \end{figure*}

                We further evaluate the robustness and generalization capability of the proposed detection framework across different urban environments. Specifically, experiments are conducted on two large-scale roadside datasets collected in Wuhan and Shanghai, which follow the same annotation protocol and evaluation metrics. Two evaluation settings are considered: in-domain evaluation, where training and testing are performed within the same city, and cross-city evaluation, where a model trained on one city is directly applied to the other without additional adaptation. This setting reflects realistic deployment scenarios in which roadside perception systems are expected to operate reliably in previously unseen cities.

                Table~\ref{tab:city_generalization_all} summarizes the multi-city detection performance of the Open-Vocabulary Fine-tuning model with a Swin-T backbone. The diagonal entries correspond to in-domain results, where strong detection accuracy is achieved in both cities. Among them, Wuhan exhibits the highest in-domain performance, reaching an mAP of 0.589 and an mAP@50 of 0.825. Such variations can be attributed to differences in scene complexity, infrastructure density, and visual appearance across urban environments.
                
                Despite noticeable differences in roadside layouts and object characteristics between the two cities, the cross-city evaluation results (off-diagonal entries in Table~\ref{tab:city_generalization_all}) demonstrate that models trained on one city maintain reasonable detection performance when transferred to the other. In most cases, the performance degradation remains moderate compared with in-domain evaluation, indicating that the detector captures transferable semantic and structural patterns rather than relying on city-specific visual cues. Moreover, the generalization behavior is relatively symmetric across the two cities, with Shanghai showing slightly stronger mutual transferability. Overall, these results indicate that the proposed detection framework exhibits solid robustness under domain shifts and is suitable for large-scale urban deployment scenarios.

                \begin{table}[!t]
                \centering
                \caption{Multi-city evaluation: in-domain accuracy and cross-city generalization.}
                \label{tab:city_generalization_all}
                \resizebox{0.9\linewidth}{!}{%
                \begin{tabular}{lcccc}
                \toprule
                \textbf{Train City} & \textbf{Test City} & \textbf{Setting} & \textbf{mAP} & \textbf{mAP@50} \\
                \midrule
                \multirow{2}{*}{Wuhan}
                & Wuhan    & In-domain   & 0.589 & 0.825 \\
                & Shanghai & Cross-city  & 0.406 & 0.651 \\
                \midrule
                \multirow{2}{*}{Shanghai}
                & Shanghai & In-domain   & 0.532 & 0.803 \\
                & Wuhan    & Cross-city  & 0.469 & 0.695 \\
                \bottomrule
                \end{tabular}%
                }
                \end{table}

                \subsection{Performance of Roadside Infrastructure Attribute Recognition}
                
                Beyond category detection, an essential feature of our framework is its ability to capture fine-grained attributes of roadside infrastructure. Rather than merely recognizing the object class, the system provides structured information on operational state, material, condition, shape, and color, enabling a more comprehensive understanding of urban roadside environments.

                As illustrated in Fig.~\ref{fig_attribute}, multiple categories of facilities are simultaneously detected and annotated with their attributes within the same scene. For traffic lights, the system distinguishes light colors (red, green, or yellow) and detects abnormal conditions such as malfunctioning bulbs or tilted poles. For traffic signs, it recognizes both prohibitory and mandatory types, further describing geometric shape (circular), dominant colors (red, blue, white), and surface integrity. Street lamps are annotated with structural and functional properties such as the number of lamp arms, illumination status, and damage condition. Ancillary facilities such as trash bins are characterized by waste type (recyclable or other waste), material (metal), color scheme, lid status, and nearby garbage accumulation.
                
                By jointly presenting object categories and attribute metadata in a structured format, the system extends perception from simple recognition to detailed condition analysis. This allows roadside management systems to assess infrastructure status, identify damaged or malfunctioning assets, and prioritize maintenance operations. The combination of accurate detection and fine-grained attribute reasoning thus ensures a more interpretable, actionable, and safety-oriented perception framework for intelligent urban roadside management.
                    
                Structured JSON Answering: the system supports a structured answering mode based on the Attribute-Based Schema. In this mode, each detected object is output in JSON format, including its category, bounding box coordinates, fine-grained attributes (e.g., color, condition, operational status), and confidence score. This structured output enables machine-readable results for downstream integration, while remaining consistent with the visual evidence.
                
\begin{figure}[!t]
                \centering
                \includegraphics[width=\columnwidth]{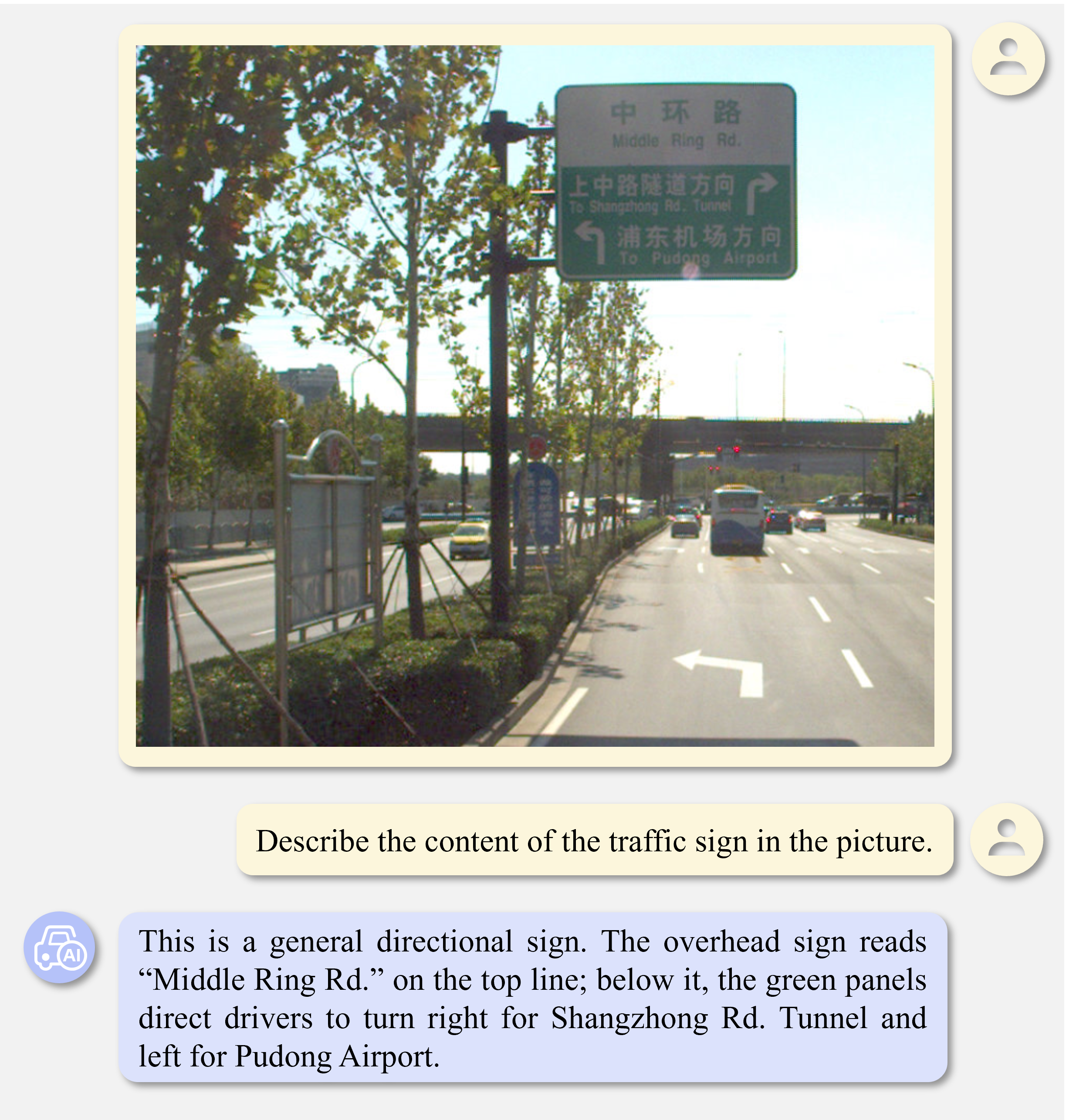}
                \caption{Qualitative example of open-ended dialogue after fine-tuning: the model answers questions about road guidance content in the scene.}
                \label{fig:dialogue_example}
                \end{figure}

                
                Beyond emitting schema-aligned JSON outputs, the fine-tuned model also supports open-ended multimodal dialogue. 
                Figure~\ref{fig:dialogue_example} shows a typical query–image pair in which the user asks about the textual and directional guidance visible on the overhead road board. 
                The model reads and returns the salient content, e.g., \emph{“Middle Ring Rd.”} and the guidance \emph{“Right To Shangzhong Rd. Tunnel; Left To Pudong Airport,”} together with the indicated turning directions shown by the arrows. 
                This example illustrates that semantics such as place names and routing instructions—information not naturally captured by attribute schemas—can be obtained through dialogue while maintaining image grounding. 
                In practice, LoRA-based adaptation and domain grounding yield concise, accurate answers without hallucinated attributes, demonstrating that the system handles free-form queries alongside structured predictions.
                
                While these qualitative results illustrate the capability of generating standardized records, ensuring the reliability of these data for automated maintenance requires quantitative verification. To rigorously evaluate the correctness of these structured JSON outputs for multimodal attribute reasoning, we define a specialized \textbf{Attribute Accuracy} metric. This metric calculates the ratio of correctly predicted attributes to the total number of attributes across all objects:
                
                \begin{equation}
                \mathrm{Accuracy} = \frac{\sum_{j=1}^{M} \sum_{k=1}^{K_j} \mathbb{I}(a_{jk}^{G} = a_{jk}^{P})}{\sum_{j=1}^{M} K_j}
                \end{equation}
                
                where $M$ denotes the number of objects, $K_j$ is the number of attributes for object $j$, $a_{jk}^G$ and $a_{jk}^P$ are the ground-truth and predicted values of attribute $k$ for object $j$, and $\mathbb{I}(\cdot)$ is the indicator function. This metric directly evaluates how well the generated attributes match the ground-truth annotations, providing a quantitative basis for the system's reasoning performance.

                Guided by this metric, we performed a detailed quantitative assessment to verify the model's precision across different facility types. To further analyze attribute recognition performance at a finer granularity, Table~\ref{tab:per_class_attribute_recognition} reports the per-class attribute accuracy of roadside infrastructure categories on the Shanghai and Wuhan datasets. The results indicate that categories like traffic lights and signs exhibit slightly lower accuracy due to the complexity of distinguishing fine-grained sub-types, such as motor vehicle versus pedestrian signals. Moreover, small-scale objects like surveillance cameras and ball bollards present greater challenges for attribute recognition, while the performance fluctuations between Shanghai and Wuhan reflect variations in scene layouts and the diverse visual appearances of infrastructure across different urban environments.

                \begin{table}[!h] 
                \centering 
                \caption{Per-class attribute recognition performance of roadside infrastructure categories on the Shanghai and Wuhan datasets after fine-tuning and dual-modality RAG.} 
                \label{tab:per_class_attribute_recognition} 
                
                \renewcommand{\arraystretch}{0.9} 
                
                \footnotesize
                
                \begin{tabular*}{\linewidth}{@{\extracolsep{\fill}} l c c } 
                    \toprule 
                    \textbf{Category} & \textbf{Shanghai Accuracy} & \textbf{Wuhan Accuracy} \\ 
                    \midrule 
                    Traffic light & 91.6 & 85.8 \\ 
                    Fire hydrant & 96.2 & 97.9 \\ 
                    Street light & 99.0 & 97.7 \\ 
                    Traffic sign & 90.1 & 95.1 \\ 
                    Bollard & 98.9 & 98.9 \\ 
                    Surveillance camera & 91.1 & 91.4 \\ 
                    Manhole & 97.7 & 97.1 \\ 
                    Trash bin & 97.4 & 97.2 \\ 
                    Ball bollard & 80.7 & 89.5 \\ 
                    Traffic cone & 97.1 & 99.6 \\ 
                    \midrule 
                    All & 95.5 & 94.1 \\ 
                    \bottomrule 
                \end{tabular*} 
                \end{table}
                
                \renewcommand{\arraystretch}{1.0}
                    
            \subsection{Comparative Experiments}
            
                \subsubsection{Object Detection Comparison}

                    \textbf{(1) Effectiveness of Fine-Tuning (Before vs. After):}
                   
\begin{figure}[!h]
                    \centering{
                    \includegraphics[width=0.48\textwidth]{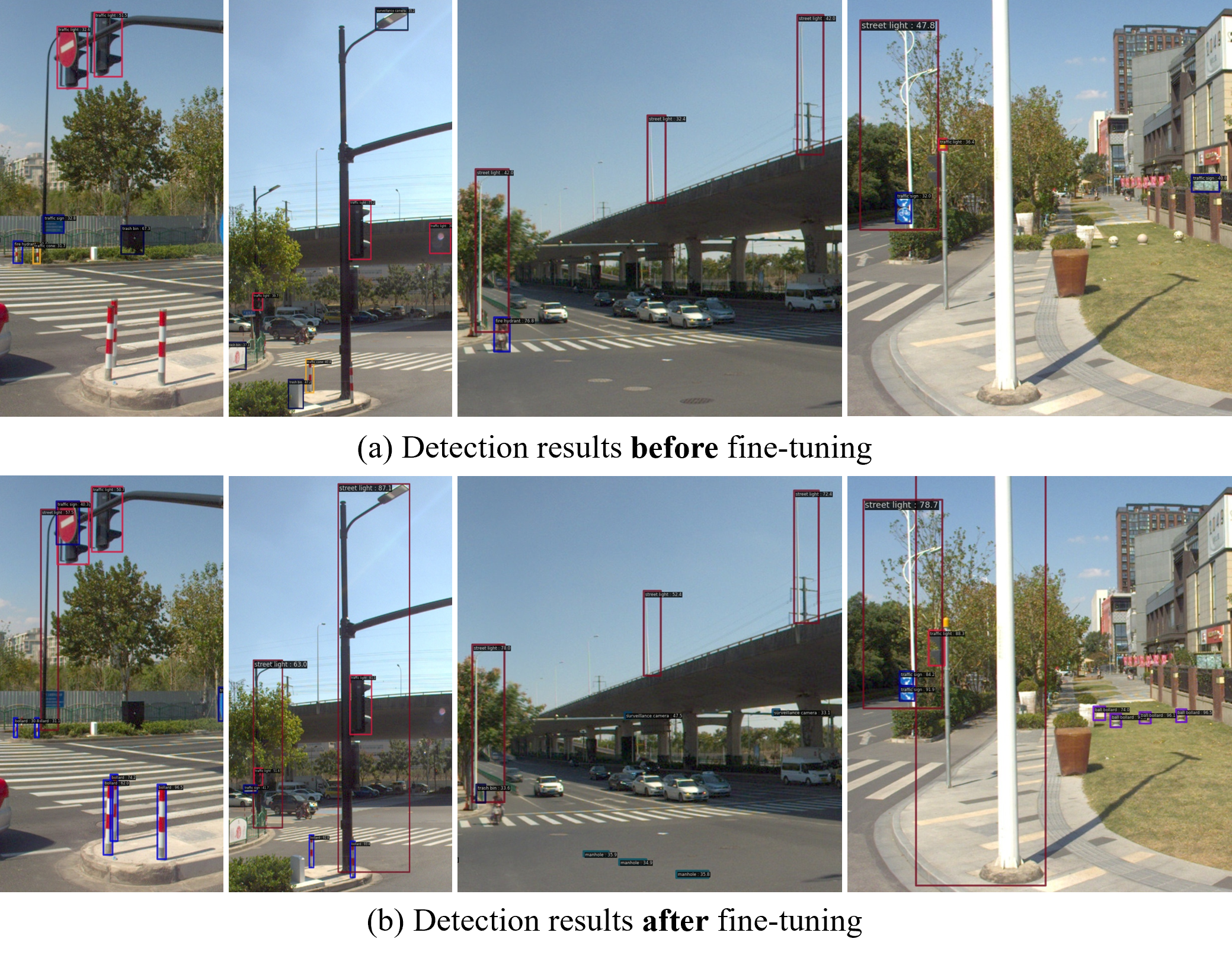}}%
                    \caption{Comparison of the after fine-tuning and before fine-tuning detection results}
                    \label{fig_detection_result_comparison}
                    \end{figure}

                    Accurate detection of roadside infrastructure serves as the foundation of the proposed framework. Fig.~\ref{fig_detection_result_comparison} presents a comparative visualization of detection results before and after fine-tuning. Fig.~\ref{fig_detection_result_comparison}(b) shows the outputs of the fine-tuned model, while Fig.~\ref{fig_detection_result_comparison}(a) depicts the baseline results. It can be clearly observed that after domain-specific fine-tuning, the model achieves more precise and complete detection, successfully identifying small, slender, and complex infrastructure elements that were previously missed or misclassified, such as streetlight poles, fire hydrants, and traffic cones. The bounding boxes are more stable and better aligned with object boundaries, reflecting enhanced localization robustness and higher category confidence. These improvements verify the effectiveness of the open-vocabulary fine-tuning strategy and confirm that the ten representative classes of roadside infrastructure can now be reliably and consistently recognized.

                    To validate these qualitative observations with quantitative evidence, we use the Wuhan dataset as a representative example to illustrate the effects of different fine-tuning strategies. As shown in Table~\ref{tab:novel_comparison}, the quantitative results reveal that fine-tuning greatly enhances the detection ability of novel categories. Specifically, the mAP@50 of traffic lights improves from 57.9 to 89.7, bollard from 1.8 to 90.7, and ball bollards from 0.0 to 80.8. These improvements demonstrate the model’s capability to capture discriminative features of small-scale and underrepresented objects after fine-tuning.
                    
                    However, certain categories remain challenging. Surveillance cameras, for instance, improve from 8.2 to 75.9 in mAP@50, but this is still lower than other categories. The primary reasons include limited training samples and small object sizes, which hinder robust detection. Nevertheless, the fine-tuned model consistently outperforms the baseline across almost all novel categories.

                    Overall, these improvements verify the effectiveness of the open-vocabulary fine-tuning strategy and confirm that the ten representative classes of roadside infrastructure can now be reliably and consistently recognized.

                    \begin{table}[!h]
                    \centering
                    \caption{Novel class detection accuracy comparison on Wuhan dataset (before vs. after fine-tuning).}
                    \label{tab:novel_comparison}
                    \resizebox{0.9\linewidth}{!}{%
                    \begin{tabular}{lcccc}
                    \toprule
                    \multirow{2}{*}{\textbf{Category}} & \multicolumn{2}{c}{\textbf{Zero-shot}} & \multicolumn{2}{c}{\textbf{Fine-tuning}} \\
                    \cmidrule(lr){2-3} \cmidrule(lr){4-5}
                     & \textbf{mAP} & \textbf{mAP@50} & \textbf{mAP} & \textbf{mAP@50} \\
                    \midrule
                    Traffic light       & 33.3 & 57.9 & 67.2 & 89.7 \\
                    Fire hydrant        & 34.8 & 61.3 & 51.4 & 80.9 \\
                    Street light        & 30.2 & 40.7 & 64.2 & 81.9 \\
                    Traffic sign        & 28.4 & 38.9 & 63.8 & 77.7 \\
                    Bollard             & 0.7  & 1.8  & 62.1 & 90.7 \\
                    Surveillance camera & 3.3  & 8.2  & 39.7 & 75.9 \\
                    Manhole             & 7.7  & 12.6 & 44.2 & 68.7 \\
                    Trash bin           & 20.9 & 30.5 & 73.8 & 91.9 \\
                    Ball bollard        & 0.0  & 0.0  & 60.0 & 80.8 \\
                    Traffic cone        & 21.1 & 37.1 & 62.5 & 87.0 \\
                    \midrule
                    All                 & 18.0 & 28.9 & 58.9 & 82.5 \\
                    \bottomrule
                    \end{tabular}%
                    }
                    \end{table}

                    \textbf{(2) Comparison with Representative Detection Models:}
                    
                    We further compare MM-Grounding DINO (MM\mbox{-}GD) with representative open\mbox{-}vocabulary detectors (GLIP, YOLO\mbox{-}World~\cite{cheng2024yoloworld}, and OV\mbox{-}DINO~\cite{wang2024ovdino}) on the roadside detection benchmark under both zero\mbox{-}shot and fine\mbox{-}tuning settings. GLIP, OV\mbox{-}DINO, and MM\mbox{-}GD use the same backbone (Swin\mbox{-}T) and the same pretraining corpora (Objects365~\cite{shao2019objects365} and GoldG), while YOLO\mbox{-}World employs YOLOv8\_L with Objects365~\cite{shao2019objects365}, GoldG, and CC\mbox{-}LiteV2 (a newly annotated 250k subset of CC3M~\cite{sharma2018conceptual}). As evidenced in Table~\ref{tab:arch_cmp}, MM\mbox{-}GD achieves the best zero\mbox{-}shot performance among the four models (Roadside mAP and mAR of 18.0 and 28.9, respectively), with clear gains in Roadside mAP@50 over GLIP (+13.5), YOLO\mbox{-}World (+6.0), and OV\mbox{-}DINO (+2.7).

                    After closed\mbox{-}set fine\mbox{-}tuning on the roadside dataset, MM\mbox{-}GD further widens the margin. In particular, Roadside mAP@50 reaches 83.1, surpassing OV\mbox{-}DINO (+4.3), YOLO\mbox{-}World (+14.0), and GLIP (+35.2). MM\mbox{-}GD also leads on Roadside mAP (59.2) and Roadside mAR (60.6). These results confirm that MM\mbox{-}GD delivers the strongest performance both in the zero\mbox{-}shot regime and after fine\mbox{-}tuning, especially on the key metrics Roadside mAP@50 and Roadside mAR.

                    \begin{table*}[!t]
                    \centering
                    \caption{Architecture comparison on the roadside benchmark under zero-shot and closed-set fine-tuning settings on Wuhan dataset.}
                    \label{tab:arch_cmp}
                    \resizebox{0.8\linewidth}{!}{%
                    \begin{tabular}{lcccccc}
                    \toprule
                    \textbf{Architecture} & \makecell{\textbf{Pre-Train Data}} & \textbf{Backbone} &
                    \makecell{\textbf{Roadside}\\\textbf{mAP@50}} &
                    \makecell{\textbf{Roadside}\\\textbf{mAP}} &
                    \makecell{\textbf{Roadside}\\\textbf{mAR}} \\
                    \midrule
                    YOLOv11 (from-scratch) & - & YOLOv11-n & 24.4 & 12.2 & 34.2 \\
                    \midrule
                    GLIP (zero-shot)     & O365, GoldG              & Swin-T    & 15.4 & 10.4  & 16.2 \\
                    YOLO-World (zero-shot) & O365+GoldG+CC-LiteV2   & YOLOv8\_L & 22.9 & 13.9  & 23.7 \\
                    OV-DINO (zero-shot)  & O365, GoldG              & Swin-T    & 26.2 & 16.7 & 28.3 \\
                    MM-GD (zero-shot)    & O365, GoldG              & Swin-T    & 28.9 & 18.0 & 28.4 \\
                    \midrule
                    GLIP (fine-tuning)   & O365, GoldG              & Swin-T    & 47.9 & 28.1 & 32.8 \\
                    YOLO-World (fine-tuning) & O365+GoldG+CC-LiteV2 & YOLOv8\_L & 69.1 & 41.7 & 49.3 \\
                    OV-DINO (fine-tuning)& O365, GoldG              & Swin-T    & 78.8 & 51.4 & 57.8 \\
                    MM-GD (fine-tuning)  & O365, GoldG              & Swin-T    & \textbf{83.1} & \textbf{59.2} & \textbf{60.6} \\
                    \bottomrule
                    \end{tabular}%
                    }
                    \end{table*}

                    To further contextualize these results, we compare from-scratch training on a recent YOLOv11 baseline against fine-tuned MM-GD on the same dataset. As shown in Table~\ref{tab:arch_cmp}, the from-scratch YOLOv11 model underperforms dramatically across Roadside mAP@50, Roadside mAP, and Roadside mAR, reflecting severe overfitting due to limited training data and weak generalization. In contrast, the fine-tuned MM-GD (Swin-T/Swin-B) attains substantially higher accuracy across all metrics. This comparison highlights that—in data-constrained roadside scenarios—adapting a strong object detector via fine-tuning is markedly more effective than training from scratch.

                    \textbf{(3) Comparative Analysis of Fine-Tuning Strategies:}
                    
                    Finally, we evaluate different fine-tuning strategies in Table~\ref{tab:strategy_summary}: closed-set fine-tuning substantially boosts performance on predefined roadside categories relative to zero-shot baselines. However, the last column also reveals a severe limitation: COCO mAP collapses to near zero after closed-set fine-tuning, indicating poor generalization outside the roadside-specific label space.
            
                    Moving beyond closed-set training, Table~\ref{tab:strategy_summary} shows that open-set continued pretraining yields a more balanced outcome: roadside mAP drops slightly relative to closed-set results, but COCO mAP improves noticeably. This suggests better transfer to generic distributions while maintaining reasonable in-domain performance.
                    
                    The open-vocabulary fine-tuning results in Table~\ref{tab:strategy_summary} demonstrate clear gains on novel roadside categories while constraining losses on base coco classes. With base/novel splits and domain prompts, the model substantially improves novel roadside detection, while base coco performance remains relatively stable. The improvement persists under a larger backbone, albeit with modest increments. Moreover, the box mAP (IoU 0.50:0.95) indicates the system retains fine localization capability on both novel and base classes, though certain base categories still leave room for improvement. 

                    It is also noteworthy that while novel categories achieve significant accuracy gains, some base categories show slight decreases in performance. This trade-off is expected, as the fine-tuning process focuses on learning new categories, which may slightly impact the generalization ability for previously known categories. The design goal of open-vocabulary fine-tuning is to strike a balance: maximizing precision for new categories while maintaining reasonable performance on base categories.
                    
                    In conclusion, closed-set training maximizes roadside accuracy but sacrifices cross-domain robustness; open-set continued pretraining enhances generic transfer with small in-domain trade-offs; open-vocabulary fine-tuning achieves the most favorable balance between precision on roadside categories and generalization to broader distributions.
                    
                    \begin{table}[!t]
                    \centering
                    \caption{Summary across fine-tuning regimes under identical evaluation on Wuhan dataset.}
                    \label{tab:strategy_summary}
                    \resizebox{\linewidth}{!}{%
                    \begin{tabular}{lcccc}
                    \toprule
                    \textbf{Fine-tuning mode} & \textbf{Backbone} &
                    \makecell{\textbf{Roadside mAP}} &
                    \makecell{\textbf{COCO mAP}} \\
                    \midrule
                    \multirow{2}{*}{Zero-shot}                    & Swin-T & 18.0 & 50.4 \\
                                                         & Swin-B & 18.5 & 52.5 \\
                    \multirow{2}{*}{Closed-set}                   & Swin-T & 57.2 & 0.1  \\
                                                         & Swin-B & 57.7 & 0.2  \\
                    \multirow{2}{*}{Open-set continued pretrain}  & Swin-T & 45.7 & 7.2  \\
                                                         & Swin-B & 46.4 & 8.3  \\
                    \multirow{2}{*}{Open vocabulary}              & Swin-T & 58.9 & 47.4 \\
                                                         & Swin-B & 59.0 & 47.6 \\
                    \bottomrule
                    \end{tabular}%
                    }
                    \end{table}

                \subsubsection{Attribute Recognition Comparision}

                    We evaluate the performance of models in generating structured JSON outputs, and Table~\ref{tab:json_performance} presents the comparison results on two city-level datasets. The results demonstrate that Qwen achieves 86.0\% accuracy on the Shanghai dataset in the zero-shot setting, reflecting its limited adaptation to roadside-specific attributes. After fine-tuning and RAG, the attribute accuracy on the Shanghai dataset improves to 95.5\%, highlighting the effectiveness of targeted domain adaptation for structured multimodal outputs.
                    
                    \begin{table}[!t]
                    \centering
                    \caption{Accuracy evaluation of JSON-based QA outputs on Shanghai (SH) and Wuhan (WH) datasets.}
                    \label{tab:json_performance}
                    \resizebox{0.45\textwidth}{!}{%
                    \begin{tabular}{lccc}
                    \toprule
                    \textbf{Method} & \textbf{Strategy} & \textbf{SH Acc} & \textbf{WH Acc} \\
                    \midrule
                    \textbf{Qwen-VL-Max} & \textbf{Fine-tuned \& RAG} & \textbf{95.5} & \textbf{94.1} \\
                    Qwen-VL-Max & Fine-tuned & 92.7 & 88.4 \\
                    Qwen-VL-Max & Zero-shot & 86.0 & 74.0 \\
                    Claude3.7 Sonnet & Zero-shot & 75.2 & 71.5 \\
                    GPT-4o & Zero-shot & 85.4 & 73.6 \\
                    Llama 4 Maverick & Zero-shot & 70.8 & 56.1 \\
                    \bottomrule
                    \end{tabular}
                    }
                    \end{table}
                   
                    Beyond detection, we compare schema-constrained JSON attribute outputs across several conversational vision–language models (VLMs). As demonstrated in Table~\ref{tab:json_performance}, the domain-adapted, schema-guided Qwen model consistently achieves the highest attribute accuracy on both city datasets, outperforming general-purpose systems on fine-grained roadside attributes. In particular, the fine-tuned Qwen-VL-Max with RAG attains an attribute accuracy of 95.5\% on the Shanghai dataset and 94.1\% on the Wuhan dataset, substantiating the benefit of domain adaptation and schema-constrained prompting for reliable structured outputs.
 
                    It is worth noting that the attribute recognition accuracy on the Wuhan dataset is slightly lower than that on the Shanghai dataset. This performance gap can be partially attributed to differences in data acquisition conditions, including weather, illumination, environmental complexity, and capture time, which introduce additional visual ambiguity and variability in real-world scenes.
                    
                    Notably, the integration of a dual-modality Retrieval-Augmented Generation (RAG) mechanism further enhances reasoning consistency and domain grounding. During inference, the \emph{textual RAG} dynamically retrieves relevant segments from professional knowledge bases, while the \emph{visual RAG} retrieves similar exemplars from the attribute-annotated image repository. The retrieved textual definitions and visual references are incorporated into the model’s multimodal context, enabling Qwen-VL to reason over both semantic and perceptual evidence. This dual retrieval mechanism ensures higher semantic fidelity, interpretability, and attribute completeness, particularly for subtle distinctions in material, functionality, or regulatory meaning that purely visual cues cannot disambiguate.
                    
                    Overall, the fine-tuned and RAG-enhanced Qwen model surpasses both zero-shot and non-knowledge-retrieval models on both city datasets, confirming the effectiveness of combining structured schema guidance with multimodal knowledge retrieval.
                    
                    In summary, the combination of schema-constrained generation with multimodal retrieval significantly improves the model’s ability to accurately recognize and interpret roadside infrastructure attributes, even for complex or ambiguous categories. The fine-tuned Qwen model, augmented with RAG, achieves robust performance across diverse urban scenarios, confirming the advantage of this approach in generating high-quality structured outputs.

        \section{Conclusions}

            This study unleashes the capabilities of Large Vision–Language Models (LVLMs) for the intelligent perception of urban roadside scenarios. By integrating an attribute-based schema with structured JSON generation and a dual-modality Retrieval-Augmented Generation (RAG) mechanism, we bridge the critical gap between unstructured visual descriptions and practical decision support. The proposed framework transforms generic perception capabilities into interpretable and machine-readable solutions, directly empowering automated infrastructure monitoring, maintenance, and operational planning.
            
            Comprehensive experiments validated the effectiveness of the proposed fine-tuning and enhancement strategies. Under open-vocabulary fine-tuning, the detector achieved 58.9~mAP on roadside categories and 47.6~mAP on COCO-style categories, demonstrating robust performance across both domain-specific and general objects. On the multimodal reasoning side, the domain-adapted Qwen-VL model—enhanced through LoRA fine-tuning and augmented with a dual-modality knowledge base constructed from professional textual references (e.g., GB~5768.2–2022) and visual exemplars—achieved 95.5\% attribute accuracy. This verifies the reliability of schema-guided and knowledge-grounded attribute interpretation.
            
            Future work will extend this framework from static image analysis to dynamic video understanding. We aim to incorporate temporal reasoning capabilities to support video-based object detection and continuous state tracking. By analyzing dynamic changes over time, the system will be better equipped to monitor the operational lifecycle of roadside assets and maintain robustness under evolving environmental conditions.

        \section {Declaration of Interests}
            The authors declare that they have no known competing financial interests or personal relationships that could have appeared to influence the work reported in this paper.

        \section {Declaration of Generative AI and AI-assisted technologies in the writing process}
            During the preparation of this work the author(s) used used different large language models in order to polish the manuscript and improve the readability. After using these tool/service, the author(s) reviewed and edited the content as needed and take(s) full responsibility for the content of the publication.
        
        \section {Acknowledgment}
            This work was jointly supported by the National Natural Science Foundation of China (No. 42130105).

\bibliography{ref}  

\begin{thebibliography}{41}
\expandafter\ifx\csname natexlab\endcsname\relax\def\natexlab#1{#1}\fi
\providecommand{\url}[1]{\texttt{#1}}
\providecommand{\href}[2]{#2}
\providecommand{\path}[1]{#1}
\providecommand{\DOIprefix}{doi:}
\providecommand{\ArXivprefix}{arXiv:}
\providecommand{\URLprefix}{URL: }
\providecommand{\Pubmedprefix}{pmid:}
\providecommand{\doi}[1]{\href{http://dx.doi.org/#1}{\path{#1}}}
\providecommand{\Pubmed}[1]{\href{pmid:#1}{\path{#1}}}
\providecommand{\bibinfo}[2]{#2}
\ifx\xfnm\relax \def\xfnm[#1]{\unskip,\space#1}\fi
\bibitem[{Ma et~al.(2022)Ma, Fan, Wang, Wu, Jiang, Xie, and Fan}]{ma2022road_cv}
\bibinfo{author}{N.~Ma}, \bibinfo{author}{J.~Fan}, \bibinfo{author}{W.~Wang}, \bibinfo{author}{J.~Wu}, \bibinfo{author}{Y.~Jiang}, \bibinfo{author}{L.~Xie}, \bibinfo{author}{R.~Fan},
\newblock \bibinfo{title}{Computer vision for road imaging and pothole detection: a state-of-the-art review of systems and algorithms},
\newblock \bibinfo{journal}{Transportation Safety and Environment} \bibinfo{volume}{4} (\bibinfo{year}{2022}) \bibinfo{pages}{tdac026}. \DOIprefix\doi{10.1093/tse/tdac026}.
\bibitem[{Bai et~al.(2022)Bai, Wu, Qi, Liu, Oguchi, and Barth}]{bai2022infrastructure}
\bibinfo{author}{Z.~Bai}, \bibinfo{author}{G.~Wu}, \bibinfo{author}{X.~Qi}, \bibinfo{author}{Y.~Liu}, \bibinfo{author}{K.~Oguchi}, \bibinfo{author}{M.~J. Barth},
\newblock \bibinfo{title}{Infrastructure-based object detection and tracking for cooperative driving automation: A survey},
\newblock in: \bibinfo{booktitle}{2022 IEEE Intelligent Vehicles Symposium (IV)}, \bibinfo{year}{2022}, pp. \bibinfo{pages}{1366--1373}. \DOIprefix\doi{10.1109/IV51971.2022.9827461}.
\bibitem[{Campbell et~al.(2019)Campbell, Both, and Sun}]{campbell2019detecting}
\bibinfo{author}{A.~Campbell}, \bibinfo{author}{A.~Both}, \bibinfo{author}{Q.~C. Sun},
\newblock \bibinfo{title}{Detecting and mapping traffic signs from google street view images using deep learning and gis},
\newblock \bibinfo{journal}{Computers, Environment and Urban Systems} \bibinfo{volume}{77} (\bibinfo{year}{2019}) \bibinfo{pages}{101350}. \URLprefix \url{https://www.sciencedirect.com/science/article/pii/S0198971519300870}. \DOIprefix\doi{https://doi.org/10.1016/j.compenvurbsys.2019.101350}.
\bibitem[{Cui et~al.(2024)Cui, Ma, Cao, Ye, Zhou, Liang, Chen, Lu, Yang, Liao et~al.}]{cui2024survey}
\bibinfo{author}{C.~Cui}, \bibinfo{author}{Y.~Ma}, \bibinfo{author}{X.~Cao}, \bibinfo{author}{W.~Ye}, \bibinfo{author}{Y.~Zhou}, \bibinfo{author}{K.~Liang}, \bibinfo{author}{J.~Chen}, \bibinfo{author}{J.~Lu}, \bibinfo{author}{Z.~Yang}, \bibinfo{author}{K.-D. Liao}, et~al.,
\newblock \bibinfo{title}{A survey on multimodal large language models for autonomous driving}  (\bibinfo{year}{2024}) \bibinfo{pages}{958--979}.
\bibitem[{Wen et~al.(2023)Wen, Yang, Fu, Wang, Cai, Li, Ma, Li, Xu, Shang et~al.}]{wen2023road}
\bibinfo{author}{L.~Wen}, \bibinfo{author}{X.~Yang}, \bibinfo{author}{D.~Fu}, \bibinfo{author}{X.~Wang}, \bibinfo{author}{P.~Cai}, \bibinfo{author}{X.~Li}, \bibinfo{author}{T.~Ma}, \bibinfo{author}{Y.~Li}, \bibinfo{author}{L.~Xu}, \bibinfo{author}{D.~Shang}, et~al.,
\newblock \bibinfo{title}{On the road with gpt-4v (ision): Early explorations of visual-language model on autonomous driving},
\newblock \bibinfo{journal}{arXiv preprint arXiv:2311.05332}  (\bibinfo{year}{2023}).
\bibitem[{Redmon et~al.(2016)Redmon, Divvala, Girshick, and Farhadi}]{redmon2016you}
\bibinfo{author}{J.~Redmon}, \bibinfo{author}{S.~Divvala}, \bibinfo{author}{R.~Girshick}, \bibinfo{author}{A.~Farhadi},
\newblock \bibinfo{title}{You only look once: Unified, real-time object detection},
\newblock in: \bibinfo{booktitle}{2016 IEEE Conference on Computer Vision and Pattern Recognition (CVPR)}, \bibinfo{year}{2016}, pp. \bibinfo{pages}{779--788}. \DOIprefix\doi{10.1109/CVPR.2016.91}.
\bibitem[{Liu et~al.(2016)Liu, Anguelov, Erhan, Szegedy, Reed, Fu, and Berg}]{liu2016ssd}
\bibinfo{author}{W.~Liu}, \bibinfo{author}{D.~Anguelov}, \bibinfo{author}{D.~Erhan}, \bibinfo{author}{C.~Szegedy}, \bibinfo{author}{S.~Reed}, \bibinfo{author}{C.-Y. Fu}, \bibinfo{author}{A.~C. Berg},
\newblock \bibinfo{title}{Ssd: Single shot multibox detector},
\newblock in: \bibinfo{booktitle}{Proceedings of the 2016 European Conference on Computer Vision (ECCV)}, \bibinfo{year}{2016}, pp. \bibinfo{pages}{21--37}. \DOIprefix\doi{10.1007/978-3-319-46448-0_2}.
\bibitem[{Liu et~al.(2025)Liu, Xie, Yuan, Liang, Dong, and Yang}]{LIU2025106377}
\bibinfo{author}{C.~Liu}, \bibinfo{author}{M.~Xie}, \bibinfo{author}{C.~Yuan}, \bibinfo{author}{F.~Liang}, \bibinfo{author}{Z.~Dong}, \bibinfo{author}{B.~Yang},
\newblock \bibinfo{title}{Training-free open-set 3d inventory of transportation infrastructure by combining point clouds and images},
\newblock \bibinfo{journal}{Automation in Construction} \bibinfo{volume}{178} (\bibinfo{year}{2025}) \bibinfo{pages}{106377}. \DOIprefix\doi{https://doi.org/10.1016/j.autcon.2025.106377}.
\bibitem[{Zhou et~al.(2022)Zhou, Han, Peng, Li, Yang, Dong, and Yang}]{ZHOU202263}
\bibinfo{author}{Y.~Zhou}, \bibinfo{author}{X.~Han}, \bibinfo{author}{M.~Peng}, \bibinfo{author}{H.~Li}, \bibinfo{author}{B.~Yang}, \bibinfo{author}{Z.~Dong}, \bibinfo{author}{B.~Yang},
\newblock \bibinfo{title}{Street-view imagery guided street furniture inventory from mobile laser scanning point clouds},
\newblock \bibinfo{journal}{ISPRS Journal of Photogrammetry and Remote Sensing} \bibinfo{volume}{189} (\bibinfo{year}{2022}) \bibinfo{pages}{63--77}. \URLprefix \url{https://www.sciencedirect.com/science/article/pii/S0924271622001265}. \DOIprefix\doi{https://doi.org/10.1016/j.isprsjprs.2022.04.023}.
\bibitem[{Han et~al.(2024)Han, Liu, Zhou, Tan, Dong, and Yang}]{HAN2024500}
\bibinfo{author}{X.~Han}, \bibinfo{author}{C.~Liu}, \bibinfo{author}{Y.~Zhou}, \bibinfo{author}{K.~Tan}, \bibinfo{author}{Z.~Dong}, \bibinfo{author}{B.~Yang},
\newblock \bibinfo{title}{Whu-urban3d: An urban scene lidar point cloud dataset for semantic instance segmentation},
\newblock \bibinfo{journal}{ISPRS Journal of Photogrammetry and Remote Sensing} \bibinfo{volume}{209} (\bibinfo{year}{2024}) \bibinfo{pages}{500--513}. \URLprefix \url{https://www.sciencedirect.com/science/article/pii/S0924271624000522}. \DOIprefix\doi{https://doi.org/10.1016/j.isprsjprs.2024.02.007}.
\bibitem[{Li et~al.(2022)Li, Zhang, Zhang, Yang, Li, Zhong, Wang, Yuan, Zhang, Hwang, Chang, and Gao}]{li2022grounded}
\bibinfo{author}{L.~H. Li}, \bibinfo{author}{P.~Zhang}, \bibinfo{author}{H.~Zhang}, \bibinfo{author}{J.~Yang}, \bibinfo{author}{C.~Li}, \bibinfo{author}{Y.~Zhong}, \bibinfo{author}{L.~Wang}, \bibinfo{author}{L.~Yuan}, \bibinfo{author}{L.~Zhang}, \bibinfo{author}{J.-N. Hwang}, \bibinfo{author}{K.-W. Chang}, \bibinfo{author}{J.~Gao},
\newblock \bibinfo{title}{Grounded language-image pre-training},
\newblock in: \bibinfo{booktitle}{2022 IEEE/CVF Conference on Computer Vision and Pattern Recognition (CVPR)}, \bibinfo{year}{2022}, pp. \bibinfo{pages}{10955--10965}. \DOIprefix\doi{10.1109/CVPR52688.2022.01069}.
\bibitem[{Liu et~al.(2023)Liu, Li, Wu, and Lee}]{liu2023improvedllava}
\bibinfo{author}{H.~Liu}, \bibinfo{author}{C.~Li}, \bibinfo{author}{Q.~Wu}, \bibinfo{author}{Y.~J. Lee},
\newblock \bibinfo{title}{Visual instruction tuning},
\newblock in: \bibinfo{editor}{A.~Oh}, \bibinfo{editor}{T.~Naumann}, \bibinfo{editor}{A.~Globerson}, \bibinfo{editor}{K.~Saenko}, \bibinfo{editor}{M.~Hardt}, \bibinfo{editor}{S.~Levine} (Eds.), \bibinfo{booktitle}{Advances in Neural Information Processing Systems}, volume~\bibinfo{volume}{36}, \bibinfo{publisher}{Curran Associates, Inc.}, \bibinfo{year}{2023}, pp. \bibinfo{pages}{34892--34916}. \URLprefix \url{https://proceedings.neurips.cc/paper_files/paper/2023/file/6dcf277ea32ce3288914faf369fe6de0-Paper-Conference.pdf}.
\bibitem[{Zhang et~al.(2023)Zhang, Li, Liu, Zhang, Su, Zhu, Ni, and Shum}]{zhang2022dino}
\bibinfo{author}{H.~Zhang}, \bibinfo{author}{F.~Li}, \bibinfo{author}{S.~Liu}, \bibinfo{author}{L.~Zhang}, \bibinfo{author}{H.~Su}, \bibinfo{author}{J.~Zhu}, \bibinfo{author}{L.~Ni}, \bibinfo{author}{H.-Y. Shum},
\newblock \bibinfo{title}{Dino: Detr with improved denoising anchor boxes for end-to-end object detection},
\newblock in: \bibinfo{booktitle}{The Eleventh International Conference on Learning Representations}, \bibinfo{year}{2023}. \URLprefix \url{https://openreview.net/forum?id=3mRwyG5one}.
\bibitem[{Radford et~al.(2021)Radford, Kim, Hallacy, Ramesh, Goh, Agarwal, Sastry, Askell, Mishkin, Clark et~al.}]{radford2021learning}
\bibinfo{author}{A.~Radford}, \bibinfo{author}{J.~W. Kim}, \bibinfo{author}{C.~Hallacy}, \bibinfo{author}{A.~Ramesh}, \bibinfo{author}{G.~Goh}, \bibinfo{author}{S.~Agarwal}, \bibinfo{author}{G.~Sastry}, \bibinfo{author}{A.~Askell}, \bibinfo{author}{P.~Mishkin}, \bibinfo{author}{J.~Clark}, et~al.,
\newblock \bibinfo{title}{Learning transferable visual models from natural language supervision},
\newblock in: \bibinfo{booktitle}{International conference on machine learning}, \bibinfo{organization}{PmLR}, \bibinfo{year}{2021}, pp. \bibinfo{pages}{8748--8763}.
\bibitem[{Cheng et~al.(2024)Cheng, Song, Ge, Liu, Wang, and Shan}]{cheng2024yoloworld}
\bibinfo{author}{T.~Cheng}, \bibinfo{author}{L.~Song}, \bibinfo{author}{Y.~Ge}, \bibinfo{author}{W.~Liu}, \bibinfo{author}{X.~Wang}, \bibinfo{author}{Y.~Shan},
\newblock \bibinfo{title}{Yolo-world: Real-time open-vocabulary object detection},
\newblock in: \bibinfo{booktitle}{2024 IEEE/CVF Conference on Computer Vision and Pattern Recognition (CVPR)}, \bibinfo{year}{2024}, pp. \bibinfo{pages}{16901--16911}. \DOIprefix\doi{10.1109/CVPR52733.2024.01599}.
\bibitem[{Behrendt et~al.(2017)Behrendt, Novak, and Botros}]{behrendt2017deep}
\bibinfo{author}{K.~Behrendt}, \bibinfo{author}{L.~Novak}, \bibinfo{author}{R.~Botros},
\newblock \bibinfo{title}{A deep learning approach to traffic lights: Detection, tracking, and classification},
\newblock in: \bibinfo{booktitle}{2017 IEEE International Conference on Robotics and Automation (ICRA)}, \bibinfo{publisher}{IEEE Press}, \bibinfo{year}{2017}, p. \bibinfo{pages}{1370–1377}. \DOIprefix\doi{10.1109/ICRA.2017.7989163}.
\bibitem[{Adedeji and Wang(2019)}]{adedeji2020intelligent}
\bibinfo{author}{O.~Adedeji}, \bibinfo{author}{Z.~Wang},
\newblock \bibinfo{title}{Intelligent waste classification system using deep learning convolutional neural network},
\newblock \bibinfo{journal}{Procedia Manufacturing} \bibinfo{volume}{35} (\bibinfo{year}{2019}) \bibinfo{pages}{607--612}. \URLprefix \url{https://www.sciencedirect.com/science/article/pii/S2351978919307231}. \DOIprefix\doi{https://doi.org/10.1016/j.promfg.2019.05.086}, \bibinfo{note}{the 2nd International Conference on Sustainable Materials Processing and Manufacturing, SMPM 2019, 8-10 March 2019, Sun City, South Africa}.
\bibitem[{Tabernik and Skočaj(2020)}]{tabernik2019deep}
\bibinfo{author}{D.~Tabernik}, \bibinfo{author}{D.~Skočaj},
\newblock \bibinfo{title}{Deep learning for large-scale traffic-sign detection and recognition},
\newblock \bibinfo{journal}{IEEE Transactions on Intelligent Transportation Systems} \bibinfo{volume}{21} (\bibinfo{year}{2020}) \bibinfo{pages}{1427--1440}. \DOIprefix\doi{10.1109/TITS.2019.2913588}.
\bibitem[{Houben et~al.(2013)Houben, Stallkamp, Salmen, Schlipsing, and Igel}]{houben2019detection}
\bibinfo{author}{S.~Houben}, \bibinfo{author}{J.~Stallkamp}, \bibinfo{author}{J.~Salmen}, \bibinfo{author}{M.~Schlipsing}, \bibinfo{author}{C.~Igel},
\newblock \bibinfo{title}{Detection of traffic signs in real-world images: The german traffic sign detection benchmark},
\newblock in: \bibinfo{booktitle}{The 2013 International Joint Conference on Neural Networks (IJCNN)}, \bibinfo{year}{2013}, pp. \bibinfo{pages}{1--8}. \DOIprefix\doi{10.1109/IJCNN.2013.6706807}.
\bibitem[{Choi and Lee(2024)}]{electronics13030615}
\bibinfo{author}{J.~Choi}, \bibinfo{author}{H.~Lee},
\newblock \bibinfo{title}{Real-time traffic light recognition with lightweight state recognition and ratio-preserving zero padding},
\newblock \bibinfo{journal}{Electronics} \bibinfo{volume}{13} (\bibinfo{year}{2024}). \URLprefix \url{https://www.mdpi.com/2079-9292/13/3/615}. \DOIprefix\doi{10.3390/electronics13030615}.
\bibitem[{Ma et~al.(2022)Ma, Fan, Wang, Wu, Jiang, Xie, and Fan}]{10.1093/tse/tdac026}
\bibinfo{author}{N.~Ma}, \bibinfo{author}{J.~Fan}, \bibinfo{author}{W.~Wang}, \bibinfo{author}{J.~Wu}, \bibinfo{author}{Y.~Jiang}, \bibinfo{author}{L.~Xie}, \bibinfo{author}{R.~Fan},
\newblock \bibinfo{title}{Computer vision for road imaging and pothole detection: a state-of-the-art review of systems and algorithms},
\newblock \bibinfo{journal}{Transportation Safety and Environment} \bibinfo{volume}{4} (\bibinfo{year}{2022}) \bibinfo{pages}{tdac026}. \URLprefix \url{https://doi.org/10.1093/tse/tdac026}. \DOIprefix\doi{10.1093/tse/tdac026}.
\bibitem[{Aygün et~al.(2024)Aygün, Kocaman, Aydemir, and Konakoğlu}]{Aygun2024Building}
\bibinfo{author}{Z.~Aygün}, \bibinfo{author}{M.~Kocaman}, \bibinfo{author}{S.~Aydemir}, \bibinfo{author}{B.~Konakoğlu},
\newblock \bibinfo{title}{Building damage detection using deep learning architecture with satellite images: The case of the 6 february 2023 kahramanmaraş earthquake},
\newblock \bibinfo{journal}{International Journal of Pioneering Technology and Engineering} \bibinfo{volume}{3} (\bibinfo{year}{2024}) \bibinfo{pages}{53--61}. \DOIprefix\doi{10.56158/jpte.2024.94.3.02}.
\bibitem[{et~al.(2024{\natexlab{a}})}]{bordes2024introduction}
\bibinfo{author}{F.~B. et~al.}, \bibinfo{title}{An introduction to vision-language modeling}, \bibinfo{year}{2024}{\natexlab{a}}. \URLprefix \url{https://arxiv.org/abs/2405.17247}. \href{http://arxiv.org/abs/2405.17247}{{\tt arXiv:2405.17247}}.
\bibitem[{et~al.(2024{\natexlab{b}})}]{achiam2023gpt4}
\bibinfo{author}{O.~et~al.}, \bibinfo{title}{Gpt-4 technical report}, \bibinfo{year}{2024}{\natexlab{b}}. \URLprefix \url{https://arxiv.org/abs/2303.08774}. \href{http://arxiv.org/abs/2303.08774}{{\tt arXiv:2303.08774}}.
\bibitem[{Liu et~al.(2023)Liu, Li, Wu, and Lee}]{liu2023visual}
\bibinfo{author}{H.~Liu}, \bibinfo{author}{C.~Li}, \bibinfo{author}{Q.~Wu}, \bibinfo{author}{Y.~J. Lee},
\newblock \bibinfo{title}{Visual instruction tuning},
\newblock in: \bibinfo{editor}{A.~Oh}, \bibinfo{editor}{T.~Naumann}, \bibinfo{editor}{A.~Globerson}, \bibinfo{editor}{K.~Saenko}, \bibinfo{editor}{M.~Hardt}, \bibinfo{editor}{S.~Levine} (Eds.), \bibinfo{booktitle}{Advances in Neural Information Processing Systems}, volume~\bibinfo{volume}{36}, \bibinfo{publisher}{Curran Associates, Inc.}, \bibinfo{year}{2023}, pp. \bibinfo{pages}{34892--34916}. \URLprefix \url{https://proceedings.neurips.cc/paper_files/paper/2023/file/6dcf277ea32ce3288914faf369fe6de0-Paper-Conference.pdf}.
\bibitem[{Bai et~al.(2023)Bai, Bai, Yang, Wang, Tan, Wang, Lin, Zhou, and Zhou}]{bai2023qwen}
\bibinfo{author}{J.~Bai}, \bibinfo{author}{S.~Bai}, \bibinfo{author}{S.~Yang}, \bibinfo{author}{S.~Wang}, \bibinfo{author}{S.~Tan}, \bibinfo{author}{P.~Wang}, \bibinfo{author}{J.~Lin}, \bibinfo{author}{C.~Zhou}, \bibinfo{author}{J.~Zhou}, \bibinfo{title}{Qwen-vl: A versatile vision-language model for understanding, localization, text reading, and beyond}, \bibinfo{year}{2023}. \URLprefix \url{https://arxiv.org/abs/2308.12966}. \href{http://arxiv.org/abs/2308.12966}{{\tt arXiv:2308.12966}}.
\bibitem[{Zhang et~al.(2024)Zhang, Fu, Liang, Zhang, Yu, Cai, and Yao}]{zhang2023trafficgpt}
\bibinfo{author}{S.~Zhang}, \bibinfo{author}{D.~Fu}, \bibinfo{author}{W.~Liang}, \bibinfo{author}{Z.~Zhang}, \bibinfo{author}{B.~Yu}, \bibinfo{author}{P.~Cai}, \bibinfo{author}{B.~Yao},
\newblock \bibinfo{title}{Trafficgpt: Viewing, processing and interacting with traffic foundation models},
\newblock \bibinfo{journal}{Transport Policy} \bibinfo{volume}{150} (\bibinfo{year}{2024}) \bibinfo{pages}{95--105}. \URLprefix \url{https://www.sciencedirect.com/science/article/pii/S0967070X24000726}. \DOIprefix\doi{https://doi.org/10.1016/j.tranpol.2024.03.006}.
\bibitem[{Liu et~al.(2024)Liu, Xue, Chen, Chen, Zhao, Wang, Hou, Li, and Peng}]{liu2024hallucination}
\bibinfo{author}{H.~Liu}, \bibinfo{author}{W.~Xue}, \bibinfo{author}{Y.~Chen}, \bibinfo{author}{D.~Chen}, \bibinfo{author}{X.~Zhao}, \bibinfo{author}{K.~Wang}, \bibinfo{author}{L.~Hou}, \bibinfo{author}{R.~Li}, \bibinfo{author}{W.~Peng}, \bibinfo{title}{A survey on hallucination in large vision-language models}, \bibinfo{year}{2024}. \URLprefix \url{https://arxiv.org/abs/2402.00253}. \href{http://arxiv.org/abs/2402.00253}{{\tt arXiv:2402.00253}}.
\bibitem[{Maaz et~al.(2024)Maaz, Rasheed, Khan, and Khan}]{Maaz2024Video}
\bibinfo{author}{M.~Maaz}, \bibinfo{author}{H.~Rasheed}, \bibinfo{author}{S.~Khan}, \bibinfo{author}{F.~Khan},
\newblock \bibinfo{title}{Video-chatgpt: Towards detailed video understanding via large vision and language models},
\newblock in: \bibinfo{booktitle}{Proceedings of the 62nd Annual Meeting of the Association for Computational Linguistics (Volume 1: Long Papers)}, \bibinfo{year}{2024}, pp. \bibinfo{pages}{12585--12602}.
\bibitem[{Chen et~al.(2024)Chen, Lin, Xu, Chai, Liang, and Wong}]{chen2024map}
\bibinfo{author}{J.~Chen}, \bibinfo{author}{B.~Lin}, \bibinfo{author}{R.~Xu}, \bibinfo{author}{Z.~Chai}, \bibinfo{author}{X.~Liang}, \bibinfo{author}{K.-Y. Wong}, \bibinfo{title}{Mapgpt: Map-guided prompting with adaptive path planning for vision-and-language navigation}, \bibinfo{year}{2024}.
\bibitem[{Bai et~al.(2023)Bai, Bai, Yang, Wang, Tan, Wang, Lin, Zhou, and Zhou}]{bai2023qwen_tech}
\bibinfo{author}{J.~Bai}, \bibinfo{author}{S.~Bai}, \bibinfo{author}{S.~Yang}, \bibinfo{author}{S.~Wang}, \bibinfo{author}{S.~Tan}, \bibinfo{author}{P.~Wang}, \bibinfo{author}{J.~Lin}, \bibinfo{author}{C.~Zhou}, \bibinfo{author}{J.~Zhou},
\newblock \bibinfo{title}{Qwen-vl: A frontier large vision-language model with versatile abilities},
\newblock \bibinfo{journal}{arXiv preprint arXiv:2308.12966} \bibinfo{volume}{1} (\bibinfo{year}{2023}) \bibinfo{pages}{3}.
\bibitem[{Hu et~al.(2022)Hu, Shen, Wallis, Allen-Zhu, Li, Wang, Wang, Chen et~al.}]{hu2021lora}
\bibinfo{author}{E.~J. Hu}, \bibinfo{author}{Y.~Shen}, \bibinfo{author}{P.~Wallis}, \bibinfo{author}{Z.~Allen-Zhu}, \bibinfo{author}{Y.~Li}, \bibinfo{author}{S.~Wang}, \bibinfo{author}{L.~Wang}, \bibinfo{author}{W.~Chen}, et~al., \bibinfo{title}{Lora: Low-rank adaptation of large language models}, \bibinfo{year}{2022}.
\bibitem[{Lewis et~al.(2020)Lewis, Perez, Piktus, Petroni, Karpukhin, Goyal, Küttler, Lewis, Yih, Rocktäschel et~al.}]{lewis2020rag}
\bibinfo{author}{P.~Lewis}, \bibinfo{author}{E.~Perez}, \bibinfo{author}{A.~Piktus}, \bibinfo{author}{F.~Petroni}, \bibinfo{author}{V.~Karpukhin}, \bibinfo{author}{N.~Goyal}, \bibinfo{author}{H.~Küttler}, \bibinfo{author}{M.~Lewis}, \bibinfo{author}{W.-t. Yih}, \bibinfo{author}{T.~Rocktäschel}, et~al.,
\newblock \bibinfo{title}{Retrieval-augmented generation for knowledge-intensive nlp tasks},
\newblock \bibinfo{journal}{Advances in Neural Information Processing Systems (NeurIPS)} \bibinfo{volume}{33} (\bibinfo{year}{2020}) \bibinfo{pages}{9459--9474}.
\bibitem[{Pan et~al.(2024)Pan, Luo, Wang, Chen, Wang, and Wu}]{pan2024vectorrag}
\bibinfo{author}{S.~Pan}, \bibinfo{author}{L.~Luo}, \bibinfo{author}{Y.~Wang}, \bibinfo{author}{C.~Chen}, \bibinfo{author}{J.~Wang}, \bibinfo{author}{X.~Wu},
\newblock \bibinfo{title}{Unifying large language models and knowledge graphs: A roadmap},
\newblock \bibinfo{journal}{IEEE Transactions on Knowledge and Data Engineering} \bibinfo{volume}{36} (\bibinfo{year}{2024}) \bibinfo{pages}{3580--3599}. \DOIprefix\doi{10.1109/TKDE.2024.3352100}.
\bibitem[{Zhou et~al.(2024)Zhou, Li, Wang, and Shen}]{zhou2024visual}
\bibinfo{author}{Y.~Zhou}, \bibinfo{author}{X.~Li}, \bibinfo{author}{Q.~Wang}, \bibinfo{author}{J.~Shen},
\newblock \bibinfo{title}{Visual in-context learning for large vision-language models},
\newblock in: \bibinfo{booktitle}{Findings of the Association for Computational Linguistics: ACL 2024}, \bibinfo{organization}{Association for Computational Linguistics}, \bibinfo{year}{2024}, pp. \bibinfo{pages}{15890--15902}. \URLprefix \url{https://aclanthology.org/2024.findings-acl.940}. \DOIprefix\doi{10.18653/v1/2024.findings-acl.940}.
\bibitem[{Cordts et~al.(2016)Cordts, Omran, Ramos, Rehfeld, Enzweiler, Benenson, Franke, Roth, and Schiele}]{cordts2016cityscapes}
\bibinfo{author}{M.~Cordts}, \bibinfo{author}{M.~Omran}, \bibinfo{author}{S.~Ramos}, \bibinfo{author}{T.~Rehfeld}, \bibinfo{author}{M.~Enzweiler}, \bibinfo{author}{R.~Benenson}, \bibinfo{author}{U.~Franke}, \bibinfo{author}{S.~Roth}, \bibinfo{author}{B.~Schiele},
\newblock \bibinfo{title}{The cityscapes dataset for semantic urban scene understanding},
\newblock in: \bibinfo{booktitle}{Proceedings of the IEEE conference on computer vision and pattern recognition}, \bibinfo{year}{2016}, pp. \bibinfo{pages}{3213--3223}.
\bibitem[{Yu et~al.(2020)Yu, Chen, Wang, Xian, Chen, Liu, Madhavan, and Darrell}]{yu2020bdd100k}
\bibinfo{author}{F.~Yu}, \bibinfo{author}{H.~Chen}, \bibinfo{author}{X.~Wang}, \bibinfo{author}{W.~Xian}, \bibinfo{author}{Y.~Chen}, \bibinfo{author}{F.~Liu}, \bibinfo{author}{V.~Madhavan}, \bibinfo{author}{T.~Darrell},
\newblock \bibinfo{title}{Bdd100k: A diverse driving dataset for heterogeneous multitask learning},
\newblock in: \bibinfo{booktitle}{Proceedings of the IEEE/CVF conference on computer vision and pattern recognition}, \bibinfo{year}{2020}, pp. \bibinfo{pages}{2636--2645}.
\bibitem[{Lin et~al.(2014)Lin, Maire, Belongie, Hays, Perona, Ramanan, Doll{\'a}r, and Zitnick}]{lin2014microsoft}
\bibinfo{author}{T.-Y. Lin}, \bibinfo{author}{M.~Maire}, \bibinfo{author}{S.~Belongie}, \bibinfo{author}{J.~Hays}, \bibinfo{author}{P.~Perona}, \bibinfo{author}{D.~Ramanan}, \bibinfo{author}{P.~Doll{\'a}r}, \bibinfo{author}{C.~L. Zitnick},
\newblock \bibinfo{title}{Microsoft coco: Common objects in context},
\newblock in: \bibinfo{booktitle}{Computer Vision -- ECCV 2014}, \bibinfo{publisher}{Springer International Publishing}, \bibinfo{address}{Cham}, \bibinfo{year}{2014}, pp. \bibinfo{pages}{740--755}.
\bibitem[{Wang et~al.(2024)Wang, Ren, Jie, Dong, Feng, Qian, Ma, Jiang, Wang, Lan et~al.}]{wang2024ovdino}
\bibinfo{author}{H.~Wang}, \bibinfo{author}{P.~Ren}, \bibinfo{author}{Z.~Jie}, \bibinfo{author}{X.~Dong}, \bibinfo{author}{C.~Feng}, \bibinfo{author}{Y.~Qian}, \bibinfo{author}{L.~Ma}, \bibinfo{author}{D.~Jiang}, \bibinfo{author}{Y.~Wang}, \bibinfo{author}{X.~Lan}, et~al.,
\newblock \bibinfo{title}{Ov-dino: Unified open-vocabulary detection with language-aware selective fusion},
\newblock \bibinfo{journal}{arXiv preprint arXiv:2407.07844}  (\bibinfo{year}{2024}).
\bibitem[{Shao et~al.(2019)Shao, Li, Zhang, Peng, Yu, Zhang, Li, and Sun}]{shao2019objects365}
\bibinfo{author}{S.~Shao}, \bibinfo{author}{Z.~Li}, \bibinfo{author}{T.~Zhang}, \bibinfo{author}{C.~Peng}, \bibinfo{author}{G.~Yu}, \bibinfo{author}{X.~Zhang}, \bibinfo{author}{J.~Li}, \bibinfo{author}{J.~Sun},
\newblock \bibinfo{title}{Objects365: A large-scale, high-quality dataset for object detection},
\newblock in: \bibinfo{booktitle}{Proceedings of the IEEE/CVF International Conference on Computer Vision (ICCV)}, \bibinfo{year}{2019}, pp. \bibinfo{pages}{8430--8439}.
\bibitem[{Sharma et~al.(2018)Sharma, Ding, Goodman, and Soricut}]{sharma2018conceptual}
\bibinfo{author}{P.~Sharma}, \bibinfo{author}{N.~Ding}, \bibinfo{author}{S.~Goodman}, \bibinfo{author}{R.~Soricut},
\newblock \bibinfo{title}{Conceptual captions: A cleaned, hypernymed, image alt-text dataset for automatic image captioning},
\newblock in: \bibinfo{booktitle}{Proceedings of the 56th Annual Meeting of the Association for Computational Linguistics (ACL)}, \bibinfo{year}{2018}, pp. \bibinfo{pages}{2556--2565}.

\end{thebibliography}

\end{document}